\documentclass[journal=dce]{CUP-JNL-DTM}%

\usepackage{graphicx}
\usepackage{multicol,multirow}
\usepackage{amsmath,amssymb,amsfonts}
\usepackage{mathrsfs}
\usepackage{amsthm}
\usepackage{rotating}
\usepackage{appendix}
\usepackage{ifpdf}
\usepackage[T1]{fontenc}
\usepackage{newtxtext}
\usepackage{newtxmath}
\usepackage{textcomp}
\usepackage[dvipsnames]{xcolor}
\usepackage{lipsum}
\usepackage[colorlinks,allcolors=blue]{hyperref}

\theoremstyle{definition}

\numberwithin{equation}{section}

\usepackage{subcaption}
\usepackage{float}
\usepackage{color} 
\usepackage{tablefootnote}
\usepackage[bottom]{footmisc}
\usepackage[super]{nth}
\usepackage{framed}
\usepackage{array}
\usepackage{booktabs}
\usepackage{verbatim}
\usepackage{eurosym}
\usepackage[all]{nowidow}
\usepackage{commath}
\usepackage{listings}
\usepackage[normalem]{ulem}
\useunder{\uline}{\ul}{}
\usepackage{todonotes}
\usepackage{bm}
\usepackage{soul}

\jname{Data/Math}
\articletype{ARTICLE TYPE}
\jyear{2025}

\begin{document}

\begin{Frontmatter}

\title[Article Title]{Adversarial Disentanglement by Backpropagation with Physics-Informed Variational Autoencoder}

\author[1]{Ioannis-Christoforos Koune}
\author[2]{Alice Cicirello}

\authormark{I. C. Koune and A. Cicirello}

\address[1]{\orgdiv{Department of Civil Engineering and Geosciences}, \orgname{Technical University of Delft}, \orgaddress{\city{Delft}, \postcode{2628 CN},  \country{Netherlands}}. \email{i.c.koune@tudelft.nl}}

\address[2]{\orgdiv{Department of Engineering}, \orgname{University of Cambridge}, \orgaddress{\city{Cambridge}, \postcode{CB2 1PZ},   \country{United Kingdom}}. \email{ac685@cam.ac.uk}}



\abstract{
Inference and prediction under partial knowledge of a physical system is challenging, particularly when multiple confounding sources influence the measured response. Explicitly accounting for these influences in physics-based models is often infeasible due to epistemic uncertainty, cost, or time constraints, resulting in models that fail to accurately describe the behavior of the system. On the other hand, data-driven machine learning models such as variational autoencoders are not guaranteed to identify a parsimonious representation. As a result, they can suffer from poor generalization performance and reconstruction accuracy in the regime of limited and noisy data. We propose a physics-informed variational autoencoder architecture that combines the interpretability of physics-based models with the flexibility of data-driven models. To promote disentanglement of the known physics and confounding influences, the latent space is partitioned into physically meaningful variables that parametrize a physics-based model, and data-driven variables that capture variability in the domain and class of the physical system. The encoder is coupled with a decoder that integrates physics-based and data-driven components, and constrained by an adversarial training objective that prevents the data-driven components from overriding the known physics, ensuring that the physics-grounded latent variables remain interpretable. We demonstrate that the model is able to disentangle features of the input signal and separate the known physics from confounding influences using supervision in the form of class and domain observables. The model is evaluated on a series of synthetic case studies relevant to engineering structures, demonstrating the feasibility of the proposed approach.} 

\end{Frontmatter}

\section*{Impact Statement}
Models of complex physical systems, such as those encountered in Structural Health Monitoring, typically fall under either the physics-based or data-driven paradigms. The former are often constrained by limited domain knowledge, while the latter can produce unrealistic predictions that are inconsistent with the known physical laws that govern the system. Hybrid approaches that integrate both physics-based and data-driven components face a trade-off between interpretability and flexibility. In variational autoencoders, which is the main focus of this paper, flexible data-driven components in the decoder can override the known physics, resulting in poor performance and loss of the physical meaning of the latent variables. This work contributes to the integration of domain knowledge with machine learning for hybrid modeling of engineering systems, by proposing an approach that aims to preserve the interpretability of physically meaningful latent variables while accounting for confounding influences in a data-driven manner.


\section{Introduction}
The aim of this work is to propose and evaluate an approach for learning disentangled representations of the underlying generative factors that characterize the behavior of an engineering system, of particular relevance for the monitoring of civil and mechanical structures. The proposed approach aims to identify and attribute variability observed in \textit{response measurements} obtained from an engineering system to variability stemming from the \textit{modeled physics}, \textit{domain}, and \textit{class} influences. We define the domain as the environmental and operational conditions that a system is exposed to, as well as other properties of the system that may not be directly specified in the model of the known physics. The class is defined as the characteristics of a structure related to the existence and extent of damage and degradation. Generally, we assume that domain information is relatively cheap and easy to collect, compared to class information. Such situations often arise when investigation by experts, costly equipment or elaborate experimental procedures are required to obtain measurements of the class variables. It is important to note that, although we view this problem from the perspective of civil and mechanical structural engineering systems, the approach described in this work can be adapted to other settings.

Our objective is to accurately infer a posterior distribution over physically meaningful latent variables, to reconstruct the structural response, quantify the associated uncertainty, and predict the damage and degradation condition of the system in previously unseen conditions. This is achieved using a limited number of noisy measurements of the structural response, domain and class variables. Due to the influence of the domain, class, and other unknown confounding factors, this will generally be an ill-posed inverse problem that requires learning a \textit{disentangled representation} \citep{Bengio2014} of the different generative factors. This task is further complicated by the limitations of physics-based models, which often represent structures under idealized nominal conditions and disregard the influence of environmental and operational variability, damage, and degradation. Most computational models of physical systems will contain simplifications and approximations due to lack of knowledge about certain aspects of the underlying physical process and to ensure computational tractability. Reducing this epistemic uncertainty is often infeasible due to cost or time constraints. As a result, only a partial description of the physical system is available in practical applications.

Generative probabilistic models such as Variational Autoencoders (VAE) \citep{Kingma2022}, Normalizing Flows \citep{Rezende2016}, and Generative Adversarial Networks (GANs) \citep{Goodfellow2014}, are a class of models that employ deep learning architectures to approximate the distribution of a given set of data and generate samples from the learned distribution. Generative probabilistic models have recently seen broader use in Structural Health Monitoring (SHM), and for constructing digital twins of structures \citep{Bacsa2025, Coraca2023, Tsialiamanis2021, Mao2021}. We propose a VAE architecture for approximating the joint distribution between the structural response and a set of physics-grounded latent variables, while accounting and correcting for the confounding influence of the domain and class of the structure, by leveraging observed domain and class variables. To achieve this, the VAE components are split into physics-based and data-driven branches, trained simultaneously in an end-to-end fashion. The data-driven branches are tasked with extracting features of the response that are informative about the domain and class variables, encoding them into the corresponding latent space, and using the latent code to augment the physics-based model predictions. Formulating the VAE as a combination of physics-based and data-driven components is not a straightforward task. The flexibility and learning capacity of feed-forward Neural Networks (NNs) that enables them to accurately model physical processes from data can be problematic when combining them with physics-based models, as the flexible NN components tend to override the known physics \citep{Takeishi2021}, resulting in inaccurate inference and overconfident or unrealistic predictions. To address this issue we propose an adversarial training objective that encourages an interpretable and parsimonious representation of the physical system by constraining the data-driven components of the VAE. Once trained, the model can be used to simultaneously perform inference over physically meaningful latent variables for new measurements as they become available, and generate samples from the predictive distribution of the response. Given a set of response measurements, the trained model can also be used to predict the corresponding domain and class variables. The proposed approach aims to:

\begin{itemize}
    \item Constrain the data-driven components of the model to avoid overriding the known physics and ground a subset of the latent variables to physically meaningful and interpretable quantities;
    \item Promote the learning of disentangled representations of the physics, domain and class generative factors, that are maximally informative about their corresponding modality while being minimally informative about other modalities.
    \item  Infer unknown non-linear relationships between features in the response measurements and additional domain and class observables that can not be directly included in the physics-based model.
    \item Improve uncertainty quantification by preventing the data-driven components of the decoder from compensating for all discrepancies between the physics-based model prediction and the measured response. 
\end{itemize}

To achieve these goals we investigate disentangled and invariant representation learning as a tool for regularizing machine learning components in VAE and properly utilizing the known physics, specified in terms of a nominal physics-based model. Additionally, we qualitatively and quantitatively evaluate the accuracy of the predictions and the complexity of the learned representation. The proposed model is assessed on three synthetic case studies and compared with fully data-driven approaches in a damage identification task. 

\section{Background}
This section aims to clarify the terminology and notation used throughout this text, summarize the necessary background, and illustrate the challenges that the proposed approach aims to address. In what follows, bold capital and lower case symbols denote matrix and vector quantities respectively. Light symbols denote scalars. Latent variables that are not directly observed and must be inferred from data are denoted as $z$, while $\phi$, $\theta$ and $\psi$ denote encoder, decoder and auxiliary regressor/classifier parameters, respectively. The symbols $x$, $c$ and $y$ denote the response, domain, and class observables, jointly referred to as the \textit{modalities} of a given physical system. When used as a subscript these symbols denote quantities that belong to a particular modality. As an example, $\bm{z}_\mathrm{y}$ denotes a set of latent variables that encode information about the class of a physical system. Throughout this text, $\mathcal{N}$ denotes the univariate or multivariate normal distribution parametrized by the mean and a scalar variance or matrix covariance respectively, and $\mathcal{U}$ denotes the uniform distribution parametrized by the lower and upper bound. The expectation of a function $f(\cdot)$ over a distribution $p(\cdot)$ is denoted as $\mathbb{E}_{p(\cdot)} \left[ f(\cdot) \right]$. Finally, a distinction is made between the underlying generative factors $s$ that determine the characteristics of the observed data, and the latent variables $z$, i.e. the learned representation of the generative factors.

\subsection{Problem setting}
Suppose that a nominal physics-based model and a dataset $\mathcal{D} = \{ (\bm{x}_{\mathrm{i}}, \bm{y}_{\mathrm{i}}, \bm{c}_{\mathrm{i}} )\}_{\mathrm{i}=1}^{N}$, composed of $N$ triplets of response measurements $\bm{x}_{\mathrm{i}}$, domain variables $\bm{c}_{\mathrm{i}}$ and class variables $\bm{y}_{\mathrm{i}}$ are available for a given system under investigation. In structural and mechanical engineering applications, the response measurements will often be displacements, strains or accelerations, measured under operating conditions, that describe the static or dynamic performance of the system. The domain variables $\bm{c}_i$ can be measurements of environmental and operational parameters, such as the location, temperature, humidity or other properties of a structure or sensor. The class variables $\bm{y}_i$ describe properties of the system that are cumbersome to obtain, such as assessments of the health condition of one or more structural components performed by experts or extracted from inspection reports. It is assumed that $\bm{y}$ is a quantity of interest to be predicted for new incoming observations of $\bm{x}$ and $\bm{c}$. Our goal is to simultaneously perform reconstruction of $\bm{x}$ and regression or classification on $\bm{y}$, and furthermore to utilize the observed domain and class variables to account for the impact of the domain and class influences on the measured response.

To highlight the intended application setting, three examples are presented in \autoref{fig:examples_overview} consisting of a beam, an oscillator, and population of bridges. In each example the available physics-based model fails to account for domain and class influences that are present in the measured response signals. In all three examples, the domain and class observables provide valuable information about the system that is necessary for accurately inferring a distribution over parameters of the physics-based model and reconstructing the response of the physical system. Furthermore, the class observables are significantly harder to measure and will not be available for future experiments. The oscillator and bridge examples also include unknown confounding influences, for which neither observations nor a physical description are available. The three examples are summarized below:

\begin{figure}[htb!]
	\centering
	\FIG{\includegraphics[width=1.0\textwidth]{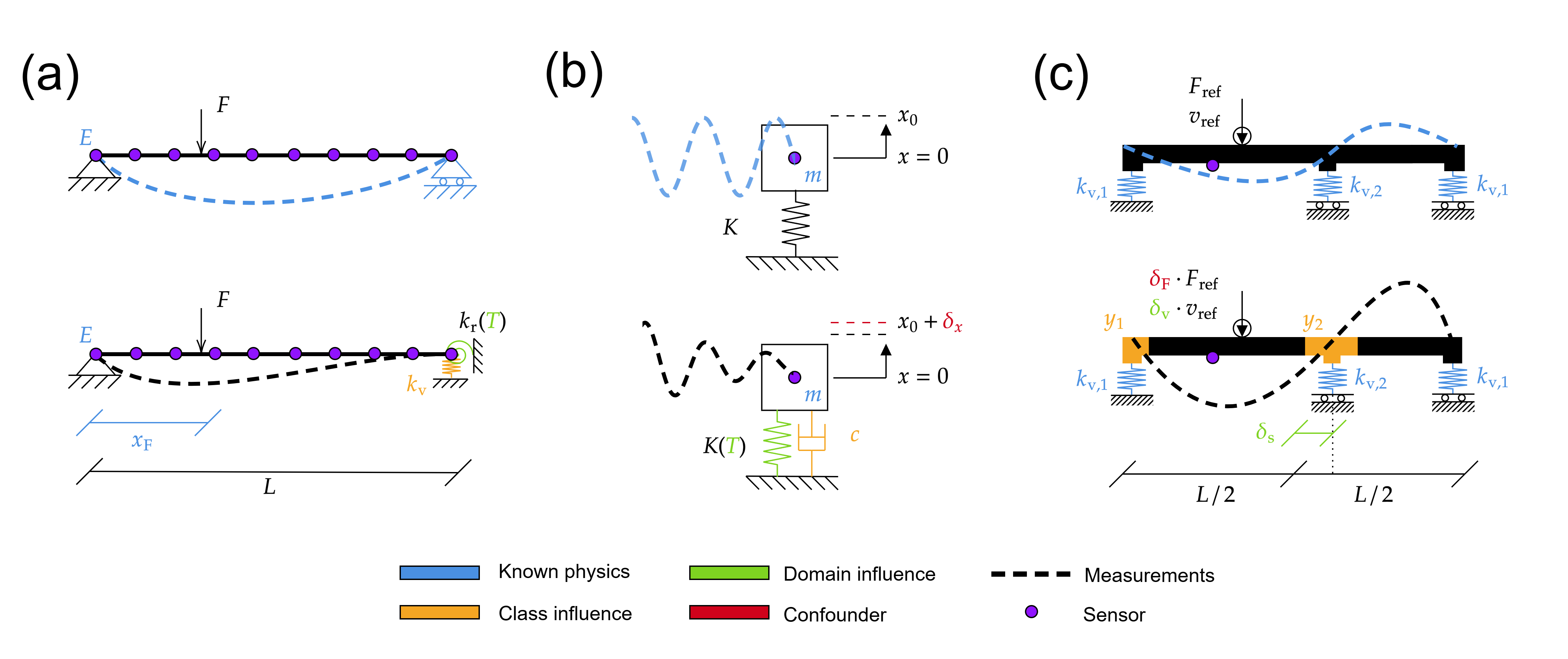}}
	\caption{Illustrative examples of the problem setting: a) Beam, b) Oscillator, and c) One member of a population of bridges. The objective is to learn components of the measured response (bottom row) that are not explicitly included in the nominal physics-based model (top row) using observations of related quantities.}
	\label{fig:examples_overview}
\end{figure}

\begin{itemize}
\item [(a)] \textbf{Beam:} In this example, the available physics-based model of a beam assumes simply supported boundary conditions and a point load acting on an unknown position. Noisy measurements of the displacement field are obtained from a set of sensors, equally spaced along the length of the beam. The domain influence is introduced as a dependence of the rotational stiffness of the right support on the temperature, and the class influence is taken as damage causing a reduction in the vertical stiffness of the right support that varies between experiments. 

\item [(b)] \textbf{Oscillator:} Noisy displacement time-series are obtained from multiple experiments, where a mass-spring-dashpot system is deflected from the equilibrium position and released to perform a harmonic oscillation. The impact of damping on the motion is neglected in the physics-based model and must be inferred from additional observations of properties of the medium. The spring stiffness is taken to depend on the ambient temperature, and there is unknown variability in the initial displaced position. The damping, temperature and variability in initial position are taken as class, domain and unknown confounding influences, respectively.

\item [(c)] \textbf{Population of bridges:} A vehicle is used to excite the response of a large set of bridges belonging to a homogeneous population, with uncertain vertical stiffness of the supports and varying position of the central pier. Each bridge is monitored by a point strain gauge that yields an influence line for the moving load. The strain gauge measurements are supplemented by qualitative assessments of the condition of the deck, obtained during inspections performed by experts and considered as class variables. The variability in the vehicle velocity and the existence of deterioration in the deck have an influence on the measured strain but are deemed too complicated to model, while the position of the pier is a known domain parameter and can be included in the physics-based model. A variability in the vehicle load is considered as an unknown confounding influence.
\end{itemize}

\subsection{Epistemic Uncertainty in Bayesian Model Updating}
Uncertainty in the modeling of physical systems can generally be classified as either aleatoric or epistemic. Aleatoric uncertainty is the component of the uncertainty due to the inherent randomness of a physical process that can not be reduced, while epistemic uncertainty stems from lack of knowledge regarding the physical process \citep{Kiureghian2009}. Epistemic uncertainty is always present to some degree in practical applications, either due to lack of knowledge or due to simplifications and approximations used to make the evaluation of the physics-based model computationally tractable. The reader is referred to \cite{Kamariotis2024} for an extended overview on classification and treatment of uncertainties in SHM applications.

Suppose that an analytical or numerical physics-based model of a structure, defined as a function $f(\bm{z}_{\mathrm{x}})$ is available. The response measurements $\bm{x}$ can then be expressed as $\bm{x} = f(\bm{z}_{\mathrm{x}}) + \epsilon$, where $\epsilon$ is a realization of a random variable quantifying the discrepancy between $f(\bm{z}_{\mathrm{x}})$ and $\bm{x}$ due to the combined influence of aleatoric and epistemic uncertainties. In the Bayesian model updating framework, the measured response of a physical system is used to update the prior knowledge, expressed in terms of a prior distribution $p(\bm{z}_\mathrm{x})$ over physically meaningful latent variables $\bm{z}_\mathrm{x}$. This is achieved by approximating the data generating process (i.e. the real-world process that generated the observations) as a combination of a deterministic physics-based model and a probabilistic model \citep{Kennedy2001}, where the latter accounts for the combined influence of epistemic and aleatoric uncertainty. In this work, it is assumed that the epistemic uncertainty stems from the confounding influence of the domain and class of a structure, and our inability (e.g. due to cost or time constraints) to account for these influences in the form and parameters of the physics-based model.

\subsection{The Variational Autoencoder}
In practical applications, the available domain knowledge is often not sufficient to guarantee that the coupled probabilistic-physical model is an accurate description of the data generating process, limiting the applicability of physics-based modeling. To remedy the lack of domain knowledge, data-driven models based on machine learning techniques have emerged as an alternative to physics-based models, where the unknown physical process is learned from measurements using flexible parametrized approximations. VAE \citep{Kingma2022, Kingma2019} are a popular data-driven approach for learning a joint distribution of data and the latent variables that are assumed to have generated the data using amortized Variational Inference (VI) \citep{Blei2017}. In VAE, the per-datapoint posterior distribution is approximated using a parametrized family of distributions, where the optimal parameters are obtained by minimizing the Kullback-Leibler divergence (KLD) between the true and approximate posteriors. The VAE is composed of an encoder network $q_{\mathrm{\bm{\phi}}}(\bm{z}|\bm{x})$ and a decoder network $p_\theta (\bm{x}|\bm{z})$, parametrized by $\phi$ and $\theta$ respectively, where $\bm{z}$ denotes latent variables that can not be observed directly and must be inferred from measurements. The encoder is typically implemented as a feed-forward NN that maps the inputs $\bm{x}$ to a conditional density over latent variables $\bm{z}$. The decoder network $p_{\mathrm{\bm{\theta}}} (\bm{x}|\bm{z})$ works in the opposite direction by approximating the density of $\bm{x}$ conditioned on $\bm{z}$. The training process for VAE consists of simultaneously optimizing the parameters of the decoder that reconstructs the observations given samples of the latent variables, and the encoder that maps inputs to a posterior distribution over these latent variables. Optimization is performed by maximizing a lower bound on the marginal likelihood of the data known as the Evidence Lower BOund (ELBO), denoted as $\mathcal{L}_{\mathrm{VAE}}$ in \autoref{eq:ELBO_VAE}. Sampling $z \sim q_\phi (\bm{z}|\bm{x})$ and evaluating the decoder yields samples from the learned distribution of the data, which in the context of civil and mechanical structural systems can be used for downstream tasks such as remaining useful life assessment.

\begin{align}
    \label{eq:ELBO_VAE}
    \begin{split}
        \mathcal{L}_{\mathrm{VAE}} (\bm{\theta}, \bm{\phi}; \bm{x}) 
        & = \mathbb{E}_{q_{\bm{\phi}}(\bm{z} | \bm{x})} \left[ \log p_{\bm{\theta}}(\bm{x} | \bm{z}) \right] - D_{\mathrm{KL}}(q_{\bm{\phi}}(\bm{z} | \bm{x}) |  | p_{\bm{\theta}}(\bm{z})) \\
        & = \log p_{\bm{\theta}}(\bm{x}) - D_{\mathrm{KL}}(q_{\bm{\phi}}(\bm{z}|\bm{x}) || p_{\bm{\theta}}(\bm{z}|\bm{x})) \\
        & \leq \log p_{\bm{\theta}}(\bm{x})
    \end{split}
\end{align}

While data-driven approaches might excel in accurately predicting the response of a physical system for a given set of input parameters when sufficient training data is available, the resulting models are typically black boxes that lack interpretability and yield no useful insights about the underlying physical process that generated the measurements. In cases where both the domain knowledge and the available data are limited, purely physics-based or data-driven approaches become infeasible, necessitating a compromise between the two extremes. Physics-enhanced machine learning (PEML) encompasses a wide range of approaches that combine machine learning with domain knowledge \citep{Cicirello2024, HaywoodAlexander2024, Cross2022, Rueden2023}. In this paradigm, the available domain knowledge can be supplemented with data, resulting in more accurate and interpretable models than would be possible with either domain knowledge or data alone. PEML approaches have the potential to reduce the required amount of data, improve accuracy and generalization performance and ensure that model predictions are consistent with the known physics. Importantly, incorporating the known physics can yield interpretable representations of physically meaningful quantities, and models that are robust and explainable. 

\subsection{Challenges in combining physics-based and data-driven components in VAE}
\label{section:challenges}
A straightforward approach to account for epistemic uncertainty in a data-driven manner would be to approximate the measured response $\bm{x}$ as the sum of the physics-based model\footnote{It is assumed that the gradients of the function with respect to the inputs can be evaluated efficiently to obtain a computationally tractable optimization problem.} $f(\bm{z}_{\mathrm{x}})$ and a trainable NN-based function $g_{\bm{\theta}}(\cdot)$, where the latter corrects the discrepancies between the physics-based model predictions and measurements. This type of hybrid model is referred to as a residual model. Parametrizing the data driven component of the residual model as $g_{\bm{\theta}}(\bm{z}_\mathrm{x})$ is not feasible when it is required that $\bm{z}_\mathrm{x}$ is interpretable: The resulting hybrid generative model has a posterior distribution $p_\mathrm{\theta}(\bm{z}_\mathrm{x} | \bm{x})$, where the latent variables $\bm{z}_\mathrm{x}$ are the input to a coupled physics-based and data-driven model, and thus no longer physically meaningful. Instead, the latent space can be partitioned as  $( \bm{z}_\mathrm{x}, \bm{z}_\mathrm{c}, \bm{z}_\mathrm{y} ) \in \bm{z}$ and the data driven component parametrized as $g_{\bm{\theta}}(\bm{z}_\mathrm{c}, \bm{z}_\mathrm{y})$, where $\bm{z}_\mathrm{c}$ and $\bm{z}_\mathrm{y}$ are physically meaningless latent variables intended to capture variability in the measured response due to the influence of the domain and the class. Assuming that the remaining aleatory uncertainties (e.g. caused by  measurement noise) are independent of the signal being measured and can be sufficiently modeled as independent and identically distributed (i.i.d.) samples of Gaussian white noise with standard deviation $\sigma_\mathrm{x}$, the response measurements can be expressed as: 

\begin{equation}
    \label{eq:responseVAE}
    \bm{x} = f(\bm{z}_{\mathrm{x}}) + g_{\bm{\theta}}(\bm{z}_\mathrm{c}, \bm{z}_\mathrm{y})+ \bm{\epsilon}_\mathrm{x},
\end{equation}

\noindent
where $\bm{\epsilon}_\mathrm{x} \sim \mathcal{N}(\bm{0},\ \sigma^2_\mathrm{x} \bm{I})$ and $\bm{I}$ is the identity matrix. Substituting $\hat{\bm{x}}_{\mathrm{p}} = f(\bm{z}_\mathrm{x})$ and $\hat{\bm{x}}_{\mathrm{d}} = g_{\bm{\theta}}(\bm{z}_\mathrm{c}, \bm{z}_\mathrm{y})$ for clarity, the resulting generative model is defined as:

\begin{equation}
    \label{eq:hybrid_decoder}
    p_\theta(\bm{x} | \bm{z}_\mathrm{x}, \bm{z}_\mathrm{c}, \bm{z}_\mathrm{y}) := \mathcal{N} \left( \hat{\bm{x}}_{\mathrm{p}} + \hat{\bm{x}}_{\mathrm{d}},\  \sigma^2_\mathrm{x} \bm{I} \right)
\end{equation}

The observables $\bm{c}$ and $\bm{y}$ can be used to ensure that the latent variables $\bm{z}_\mathrm{c}$ and $\bm{z}_\mathrm{y}$ encode information about the domain and class of the structure by simultaneously training auxiliary tasks $r_\mathrm{c}(\bm{c} | \bm{z}_\mathrm{c})$ and $r_\mathrm{y}(\bm{y} | \bm{z}_\mathrm{y})$. The resulting hybrid generative model $p_{\bm{\theta}}(\bm{x} | \bm{z}_\mathrm{x}, \bm{z}_\mathrm{c}, \bm{z}_\mathrm{y})$ can be coupled with variational posteriors $q_{\mathrm{\phi}_{\mathrm{x}}}(\bm{z}_\mathrm{x} | \bm{x}) q_{\mathrm{\phi}_{\mathrm{c}}}(\bm{z}_\mathrm{c} | \bm{x}) q_{\mathrm{\phi}_{\mathrm{y}}}(\bm{z}_\mathrm{y} | \bm{x})$ to yield an architecture similar to VAE. This is the architecture derived in Section \ref{section:approach}, without the additional constraints. It should be noted that the assumption of an additive structure for the discrepancy term $g_{\bm{\theta}}(\bm{z}_\mathrm{c}, \bm{z}_\mathrm{y})$ and the uncertainty term $\bm{\epsilon}_\mathrm{x}$ presented in \autoref{eq:responseVAE} is suitable for many physical systems and is commonly employed in hybrid models \citep{Cross2022}. We use it without loss of generality with the aim of promoting clarity and interpretability. Depending on domain knowledge regarding the problem at hand, a multiplicative or other form can also be specified. The implications of the additivity assumption are further discussed in Sections \ref{section:uncertainty_quantification} and \ref{section:assumptions_and_limitations}.

Neither the additive structure of the hybrid physics-based and data-driven model, nor the specified parametrization of the residual term $g_{\bm{\theta}}(\bm{z}_{\mathrm{c}}, \bm{z}_{\mathrm{y}})$ ensure that the known physics will be utilized by the model, or that $\bm{z}_\mathrm{x}$ will be physically meaningful. Without further constraints, the model can learn combinations of arbitrary predictions from the physics-based and data-driven components $f(\bm{z}_\mathrm{x})$ and $g_{\bm{\theta}}(\bm{z}_{\mathrm{c}}, \bm{z}_{\mathrm{y}})$ that sum to an accurate prediction. To see why, it is sufficient to consider the form of the objective given in \autoref{eq:ELBO_VAE}. Both the encoder and decoder are aligned in the task of maximizing the reconstruction term $\mathbb{E}_{q_{\bm{\phi}}(\bm{z} | \bm{x})} \left[ \log p_{\bm{\theta}}(\bm{x} | \bm{z}) \right]$, and the data-driven components of the model will account for discrepancies between the physics-based model prediction and measurements up to some level of noise. This results in an entangled representation, where the data-driven components override the known physics and the physics-grounded latent variables loose their physical meaning. 

\begin{figure}[htb!]
	\centering
	\FIG{\includegraphics[width=1.0\textwidth]{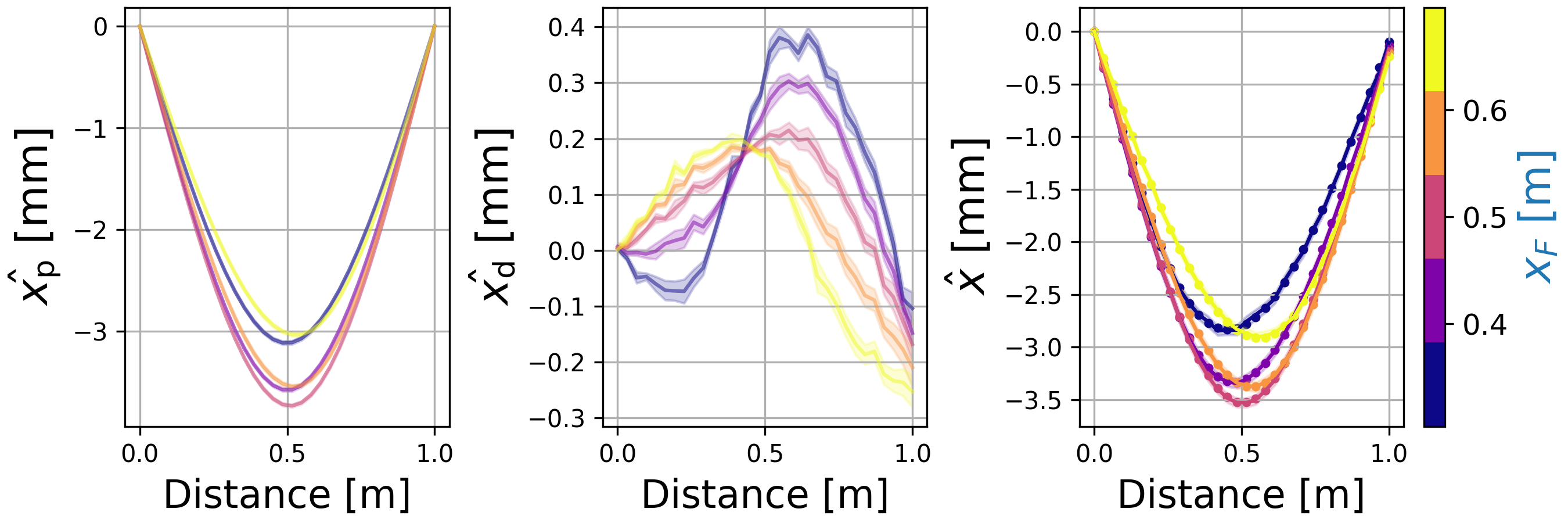}}
	\caption{Demonstration of the data-driven component of the decoder $g_{\bm{\theta}}(\bm{z}_\mathrm{c}, \bm{z}_\mathrm{y})$ overriding the physics-based model $f(\bm{z}_\mathrm{x})$. The effect of varying the position of the load $x_\mathrm{F}$ should be described by the known physics, but is instead captured by the data-driven components.}
	\label{fig:introduction_issue_1}
\end{figure}

This issue is illustrated in \autoref{fig:introduction_issue_1} using the beam case study shown in \autoref{fig:examples_overview}(a). Further details of the case study are provided in Section \ref{section:beam_example}. In this example, a VAE trained on a dataset $\mathcal{D}= \{ (\bm{x}_i, \bm{c}_i, \bm{y}_i ) \}_{i=1}^N$ is evaluated on a new set of input measurements $\bm{x}$, generated from the ground truth data generating process by linearly varying the position of the load $x_\mathrm{F}$. It can be seen that the effect of this variation on the measured response is  largely captured by the data-driven component of the decoder, which overrides the known physics, despite the fact that the physics-based model includes the load position as an input parameter. The extent to which the data-driven components override the known physics can be inconsistent and hard to predict, and will depend on the neural network architectures and the physics of the problem at hand.

The results presented in \autoref{fig:introduction_issue_1} are obtained under the assumption of a factorized variational posterior. Ideally, a single encoder with shared parameters $q_{\mathrm{\phi}}(\bm{z}_\mathrm{x}, \bm{z}_\mathrm{c}, \bm{z}_\mathrm{y} | \bm{x})$ would be used for the three subsets of the latent space $\bm{z}_\mathrm{x}, \bm{z}_\mathrm{c}$ and $\bm{z}_\mathrm{y}$. However, using a shared encoder leads to further degradation of the performance as noted by \cite{Ilse2020}. This example demonstrates how the interaction between the physics-based and data-driven components, which determines the resulting learned representation, depends on the capacity and flexibility of the individual machine learning components. Standard VAE offer no mechanism to control this interaction, and are therefore unable to guarantee that the physics will be utilized correctly in the presence of flexible NN-based decoder components.

\section{Proposed Approach}
\label{section:approach}
To address the issues presented in Section \ref{section:challenges}, we propose an approach that takes advantage of the domain and class observables to constrain the approximate posterior distribution. Our approach ensures that each subset of the latent variables only encodes information that is relevant to the corresponding modality. This constraint in turn limits the amount of information available to the data-driven components of the decoder, preventing them from correcting every discrepancy between the physics-based model and measurements, and from overriding the known physics. This is achieved by imposing a latent bottleneck structure to the model, combined with an adversarial training objective. A detailed description of the model architecture and derivation of the training objective are provided in Section \ref{section:approach_details}, followed by a brief discussion in Section \ref{section:approach_discussion}. A method for quantitatively assessing the information encoded in subsets of the latent variables is presented in Section \ref{section:quantitative_evaluation}.

\subsection{Detailed description of the model}
\label{section:approach_details}
\begin{figure}[htb!]
    \centering
    \FIG{\includegraphics[width=1.0\textwidth]{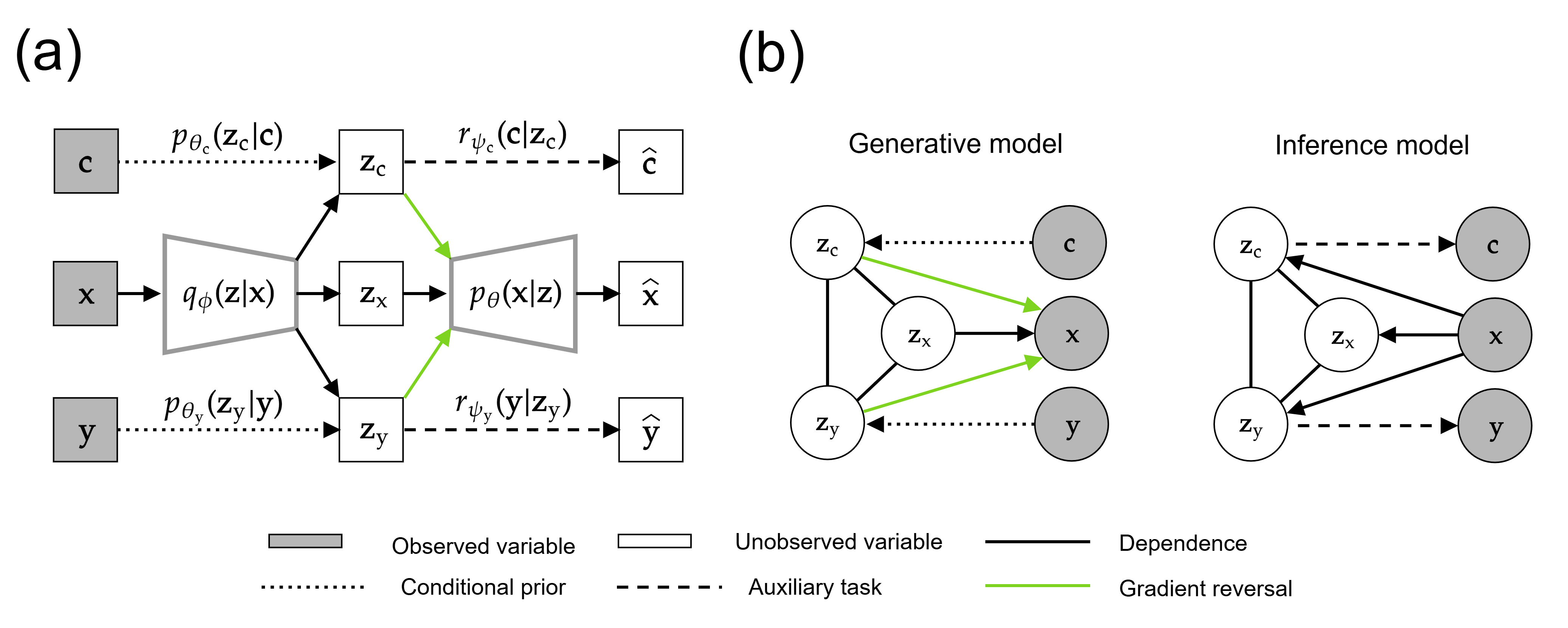}}
    \caption{a) Schematic diagram illustrating the components of the model and the encoder-decoder architecture, and b) Detailed structure of the dependencies in the generative and inference models.}
    \label{fig:dpivae_schematic}
\end{figure}

It is assumed that three generative factors, the underlying physics of the structure, the domain, and the class, contribute to the measured response $\bm{x}$. Conversely, the latent variables are partitioned into subsets\footnote{It is emphasized that the separation of the latent variables is only semantic and used for clarity. In practice, they can be the output of a single encoder with shared parameters.} $( \bm{z}_\mathrm{x}, \bm{z}_\mathrm{c}, \bm{z}_\mathrm{y} ) \in \bm{z}$. The latent variables are the input to the hybrid probabilistic decoder $p_\theta(\bm{x} | \bm{z})$, wherein a NN-based function $g_{\bm{\theta}}(\bm{z}_\mathrm{c},\bm{z}_\mathrm{y})$ accounts for discrepancies between the measured response $\bm{x}$ and physics-based model prediction $f(\bm{z}_\mathrm{x})$. To ensure that information relevant to the domain and class is encoded in the corresponding subsets $\bm{z}_\mathrm{c}$ and $\bm{z}_\mathrm{y}$, we utilize two auxiliary decoders $r_{\psi_{\mathrm{c}}}(\bm{c} | \bm{z}_c)$ and $r_{\psi_{\mathrm{y}}}(\bm{y} | \bm{z}_y)$. The latent variables $\bm{z}_\mathrm{c}$ and $\bm{z}_\mathrm{y}$ are assigned conditional prior distributions $p_{\theta_\mathrm{c}}(\bm{z}_\mathrm{c} | \bm{c})$ and $p_{\theta_{\mathrm{y}}}(\bm{z}_\mathrm{y} | \bm{y})$ respectively, while the physics-grounded latent variables $\bm{z}_\mathrm{x}$ are assigned a distribution $p(\bm{z}_\mathrm{x})$ based on the available prior knowledge. A schematic illustration of the architecture is provided in \autoref{fig:dpivae_schematic}(a).

To minimize the reconstruction error, the encoder tends to maximize the information in the posterior distribution over $\bm{z}$ that can be used to predict $\bm{x}$, $\bm{c}$ and $\bm{y}$, subject to the regularization imposed by the prior distribution. For $\bm{z}_\mathrm{c}$ and $\bm{z}_\mathrm{y}$, this information includes features from the input signal $\bm{x}$ that are predictive of $\bm{c}$ and $\bm{y}$, but also irrelevant features that are only predictive of $\bm{x}$. These features can include systematic errors stemming from partial knowledge of the physics, and the influence of unknown confounding factors in the measurements. This \textit{superfluous information} \citep{Federici2020}, i.e. information in $\bm{z}_\mathrm{c}$ and $\bm{z}_\mathrm{y}$ that is not predictive of $\bm{c}$ and $\bm{y}$, can enable the data-driven component of the decoder to override the known physics. Motivated by this observation we aim to simultaneously maximize the information in $\bm{z}_\mathrm{c}$ and $\bm{z}_\mathrm{y}$ that is predictive of $\bm{c}$ and $\bm{y}$, while minimizing the information that is predictive of $\bm{x}$. This trade-off can be formalized in terms of the Mutual Information (MI), a measure of the dependence between two random variables \citep{Cover2006}. Denoting the MI between $\bm{x}$ and $\bm{z}$ for an encoder parametrized by $\bm{\phi}$ as $I_\mathrm{\phi}(\bm{x};\bm{z})$, and introducing the trade-off parameters $\lambda_\mathrm{c}, \lambda_\mathrm{y}$, we define the following relaxed Lagrangian objectives: 

\begin{align}
    \label{eq:vib_objective}
    \begin{split}
    \mathcal{L}_\mathrm{y}(\bm{\phi}; \lambda_\mathrm{y}) & =  \ I_\mathrm{\phi}(\bm{y};\bm{z}_\mathrm{y}) - \lambda_\mathrm{y} I_\mathrm{\phi}(\bm{x}; \bm{z}_\mathrm{y}) \\
    \mathcal{L}_\mathrm{c}(\bm{\phi}; \lambda_\mathrm{c}) & =  \ I_\mathrm{\phi}(\bm{c};\bm{z}_\mathrm{c}) - \lambda_\mathrm{c} I_\mathrm{\phi}(\bm{x}; \bm{z}_\mathrm{c}) \\
    \end{split}
\end{align}

The quantities described in \autoref{eq:vib_objective} are optimized indirectly through a latent bottleneck structure \citep{Tishby2000, Alemi2019,Fischer2020, Moyer2018} combined with adversarial training, as described in the following informal sketch. Note that in the inference model shown in \autoref{fig:dpivae_schematic}(b), the latent variables $\bm{z}_\mathrm{y}$ do not depend directly on $\bm{y}$ (and analogously for $\bm{c}$). This is the conditional independence assumption typically used in the Information Bottleneck framework. Intuitively, the encoder is forced to distill the relevant information in $\bm{x}$ that is necessary for reconstructing $\bm{c}$ and $\bm{y}$ into the latent variables $\bm{z}_\mathrm{c}$ and $\bm{z}_\mathrm{y}$. This results in the maximization of the $I_\mathrm{\phi}(\bm{c};\bm{z}_\mathrm{c})$ and $I_\mathrm{\phi}(\bm{y};\bm{z}_\mathrm{y})$ terms in \autoref{eq:vib_objective} during training. The additional requirement of minimizing $I_\mathrm{\phi}(\bm{x};\bm{z}_\mathrm{c})$ and $I_\mathrm{\phi}(\bm{x};\bm{z}_\mathrm{y})$ can be satisfied by introducing a Gradient Reversal Layer (GRL) \citep{Ganin2015} at the input of the data-driven decoder component $g_{\bm{\theta}}(\bm{z}_{\mathrm{c}}, \bm{z}_{\mathrm{y}})$. During optimization, the gradient signal propagated backwards from $g_{\bm{\theta}}(\bm{z}_{\mathrm{c}}, \bm{z}_{\mathrm{y}})$ to the encoder is scaled by $-\lambda$, while the forward pass remains unchanged. Therefore, the GRL can be thought of as a pseudo function $R_\mathrm{\lambda}(\bm{z})$ such that $R_\mathrm{\lambda}(\bm{z}) = \bm{z}$ and $\frac{\mathrm{d}R_\mathrm{\lambda}}{\mathrm{d} \bm{z}} = -\lambda \bm{I}$. Positive values of $\lambda$ correspond to adversarial training. Conversely, negative values make the training ``collaborative''. The absolute value of $\lambda$ determines the strength of the adversarial or collaborative objective, with larger values corresponding to a stronger regularization effect. By turning the decoder $p_{\mathrm{\theta}}(\bm{x}|\bm{z}_\mathrm{x}, \bm{z}_\mathrm{c}, \bm{z}_\mathrm{y})$ into an adversary, the GRL penalizes information in $\bm{z}_\mathrm{c}$ and $\bm{z}_\mathrm{y}$ that contributes to the reconstruction of $\bm{x}$, biasing the encoder towards representations that are minimally informative about $\bm{x}$.

The full structure of the model is shown in \autoref{fig:dpivae_schematic}(b). The variational lower bound can be obtained by considering the marginal likelihood over observed variables as shown in \autoref{eq:marginal_likelihood}.

\begin{align}
    \label{eq:marginal_likelihood}
    \begin{split}
    \mathcal{L}(\bm{\theta}, \bm{\phi}; \bm{x}, \bm{c}, \bm{y}) & = \mathbb{E}_{q_\mathrm{\phi}(\bm{z}_\mathrm{x},\bm{z}_\mathrm{c},\bm{z}_\mathrm{y} | \bm{x})} \left[ \log \frac{p_\mathrm{\theta}(\bm{x},\bm{c},\bm{y},\bm{z}_\mathrm{x}, \bm{z}_\mathrm{c}, \bm{z}_\mathrm{y})}{q_{\mathrm{\phi}}(\bm{z}_\mathrm{x}, \bm{z}_\mathrm{c}, \bm{z}_\mathrm{y} | \bm{x})} \right] \\
    & \leq \log p_\mathrm{\theta}(\bm{x},\bm{c},\bm{y})
    \end{split}
\end{align}

\noindent
Rearranging the terms in \autoref{eq:marginal_likelihood}, noting that the generative model factorizes as $p(\bm{x}, \bm{c}, \bm{y}, \bm{z}_\mathrm{x}, \bm{z}_\mathrm{c}, \bm{z}_\mathrm{y}) = p_\mathrm{\theta}(\bm{x} | \bm{z}_\mathrm{x}, \bm{z}_\mathrm{c}, \bm{z}_\mathrm{y}) p(\bm{z}_\mathrm{x}) p_{\mathrm{\theta}_{\mathrm{c}}}(\bm{z}_\mathrm{c} | \bm{c}) p_{\mathrm{\theta}_{\mathrm{y}}}(\bm{z}_\mathrm{y} | \bm{y}) p(\bm{c}) p(\bm{y})$, yields the following expression for the lower bound:

\begin{align}
    \label{eq:lower_bound}
    \begin{split}
    \mathcal{L}(\bm{\theta}, \bm{\phi}; \bm{x},\bm{c},\bm{y}) & = \ \mathbb{E}_{q_\mathrm{\phi}(\bm{z}_\mathrm{x},\bm{z}_\mathrm{c},\bm{z}_\mathrm{y} | \bm{x})} \left[ \log p_\mathrm{\theta}(\bm{x} | \bm{z}_\mathrm{x}, \bm{z}_\mathrm{c}, \bm{z}_\mathrm{y}) \right] \\
    & - D_\mathrm{KL} \left( q_\mathrm{\phi}(\bm{z}_\mathrm{x},\bm{z}_\mathrm{c},\bm{z}_\mathrm{y} | \bm{x}) \ || \ p(\bm{z}_\mathrm{x}) p_{\mathrm{\theta}_{\mathrm{c}}}(\bm{z}_\mathrm{c} | \bm{c}) p_{\mathrm{\theta}_{\mathrm{y}}}(\bm{z}_\mathrm{y} | \bm{y}) \right) \\
    \end{split}
\end{align}

\noindent
Including the auxiliary tasks and additional regularization hyperparameters commonly used in representation learning, we rewrite the loss function as:

\begin{equation}
    \label{eq:objective_final}
    \begin{split}
    \mathcal{L}(\bm{\theta}, \bm{\phi}, \bm{\psi}; \bm{x}, \bm{c}, \bm{y}) & = \mathbb{E}_{q_{\mathrm{\phi}(\bm{z}_{\mathrm{x}}, \bm{z}_{\mathrm{c}}, \bm{z}_{\mathrm{y}} | \bm{x})}} \left[ \alpha_\mathrm{x}   \log p_{\mathrm{\theta}}(\bm{x} | \bm{z}_{\mathrm{x}}, \bm{z}_{\mathrm{c}}, \bm{z}_{\mathrm{y}}) + \alpha_\mathrm{c}   \log r_{\mathrm{\psi}_{\mathrm{c}}}(\bm{c} | \bm{z}_{\mathrm{c}}) + \alpha_\mathrm{y}   \log r_{\mathrm{\psi}_{\mathrm{y}}}(\bm{y} | \bm{z}_{\mathrm{y}}) \right] \\
    & - \beta   D_{\mathrm{KL}} \left( q_{\mathrm{\phi}} (\bm{z}_{\mathrm{x}}, \bm{z}_{\mathrm{c}},\bm{z}_{\mathrm{y}} | \bm{x}) || p(\bm{z}_\mathrm{x})   p_{\mathrm{\theta}_{\mathrm{c}}}(\bm{z}_{\mathrm{c}}  |  \bm{c})   p_{\mathrm{\theta}_{\mathrm{y}}}(\bm{z}_{\mathrm{y}} | \bm{y}) \right) \\
    \end{split}
\end{equation}

The loss function described in \autoref{eq:objective_final} includes additional regularization hyperparameters that can be used to balance the contribution of different terms. These are included for completeness, and are not used in the experiments described in Section \ref{section:case_studies}. A scaling factor $\beta > 0$ on the KLD is commonly included in the ELBO as a means of adjusting the strength of the regularization imposed by the KLD term, and to control the capacity of the probabilistic encoder. It is often beneficial to begin training with $\beta = 0$ and gradually increase it to $\beta = 1$ using an annealing scheme such as the one proposed by \cite{Bowman2016}. Annealing $\beta$ can prevent the model from getting stuck in local minima of the KLD, and the posterior distribution from degenerating to the prior distribution. Conversely, setting $\beta > 1$ can promote unsupervised disentanglement \citep{Higgins2016}. The impact of $\beta$ is extensively discussed in the relevant literature, provided in Section \ref{section:previous_work}. Additionally, the log-likelihood function $\log p_{\mathrm{\theta}}(\bm{x} | \bm{z}_{\mathrm{x}}, \bm{z}_{\mathrm{c}}, \bm{z}_{\mathrm{y}})$, and auxiliary decoders $\log r_{\mathrm{\psi}_\mathrm{c}}(\bm{c} | \bm{z}_{\mathrm{c}})$ and $\log r_{\mathrm{\psi}_\mathrm{y}}(\bm{y} | \bm{z}_{\mathrm{y}})$ are assigned weights $\alpha_\mathrm{x}$, $\alpha_\mathrm{c}$ and $\alpha_\mathrm{y}$ respectively to allow for balancing the relative strength of these terms (see e.g. \citep{Joy2022, Ilse2020, Sun2022}). It is important to note that using values other than unity for $\alpha_\mathrm{x}, \alpha_\mathrm{c}, \alpha_\mathrm{y}$ and $\beta$ can have a significant impact on the interpretation of the ELBO and the inferred posterior distribution. Details of the implementation can be found in Appendix \ref{appendix:appendix_A}.


\subsection{Discussion of the approach}
\label{section:approach_discussion}

The latent bottleneck structure and GRL have several important implications for inference, conditional generation, and uncertainty quantification. These are discussed here, and demonstrated through the case studies presented in Section \ref{section:case_studies}. 

\subsubsection{Interpretability}
The main objective of the approach is to ensure that the known physics are properly utilized, which we interpret as variability in the generative factors being preferentially captured by the physics-grounded subset of the latent variables. As a result, the posterior distribution of the physics-grounded latent variables will be influenced by the domain and class contributions to the measured response, allowing for domain and class influences to be interpreted in terms of their effect on the physics-grounded latent variables. Therefore, the posterior over physics-grounded latent variables might not necessarily be accurate in the sense of a point estimate of a physical quantity obtained from the posterior being close to the underlying ``true value''. This is also in part due to the data-driven component of the decoder, which can yield a constant but not necessarily zero prediction when domain or class influences are not present in the measured response.

\subsubsection{Conditional generation}
The use of conditional prior networks and separate branches for the domain and class modalities makes it possible to perform data imputation and conditional generation. When conditioning on domain or class variables, the accuracy of the generated response will depend on the degree to which the corresponding influence is accounted for by the known physics. If the influence is primarily accounted for by the physics, the predicted response might become insensitive to changes in the domain or class latent variables. In this case, more accurate conditional generation might be possible by fixing the values of the physics-grounded latent variables based on domain knowledge. Another implication of the architecture is that only measurements of the response are needed to evaluate the model. Throughout this work, the model is evaluated only on response measurements, without using the domain observables. This did not result in any noticeable difference in accuracy, compared to using the domain observables.

\subsubsection{Uncertainty quantification}
\label{section:uncertainty_quantification}
The uncertainty associated with the predicted response stems from the approximate posterior distribution and the probabilistic decoder, and represents the combined influence of aleatoric and epistemic uncertainties. Without the GRL, and given sufficient data, the model would compensate for systematic discrepancies between the physics-based model and response measurements in a data-driven manner. In this case, the uncertainty in the reconstructed response would only represent aleatoric uncertainty. In contrast, if no data-driven component is used in the decoder, the uncertainty would also include epistemic uncertainty due to domain and class influences that are not included in the physics-based model. In our approach, part of this epistemic uncertainty is accounted for in a data-driven manner. Therefore, the proposed approach is expected to yield uncertainty bounds somewhere in-between these two extremes. We emphasize that the estimated uncertainty does not include the uncertainty over model parameters $\bm{\theta}$, $\bm{\phi}$ and $\bm{\psi}$. Finally, it is important to consider that the additivity assumption in \autoref{eq:hybrid_decoder} is unlikely to hold for many physical systems. The model is expected to perform sub-optimally in such cases, resulting in inaccurate uncertainty estimates. None of the case studies presented in \ref{section:case_studies} satisfy the additivity assumption, demonstrating that the model can still be feasibly applied in such cases.

\subsubsection{Formulation of the latent space}
The choice of a continuous latent representation for the domain and class variables provides a number of advantages over directly representing the variables themselves, i.e. using the same number and type (e.g. categorical) of latent variables as the domain and class variables. The mapping to a low-dimensional continuous latent space enables the model to deal with high-dimensional domain and class variables, and can improve generalization by promoting the encoding of richer representations of the domain and class \citep{Joy2022}. From an implementation perspective, it is convenient if the decoder inputs are not dependent on the type and dimensionality of the domain and class variables. Furthermore, for discrete and categorical domain and class variables, the latent space enables the model to make a continuous approximation by interpolating over the continuous latent space. Broadly speaking, the continuous latent space allows for more flexibility in the representation of the domain and class variables. Finally, the lack of an independence assumption facilitates the use of a single probabilistic encoder, potentially reducing the amount of trainable parameters in the model and allowing for more complex and expressive encoder formulations. In our experiments we did not observe a decrease in performance when using a single encoder for all latent variables, compared to using separate encoders for each subset of the latent space, when combined with adversarial training.

\subsection{Description of the quantitative assessment approach}
\label{section:quantitative_evaluation}

Quantitatively assessing disentanglement is a challenging problem and several metrics have been proposed \citep{Higgins2016, Chen2018, Kim2018}. This difficulty can be partially attributed to the lack of a consistent definition of disentanglement \citep{Locatello2019}. In practice, the degree of disentanglement achieved by a model is often evaluated based on subjective expectations stemming from domain knowledge \citep{Vowels2019}. Our proposed approach aims to achieve ``one-way'' disentanglement and is conceptually more akin to techniques that temper the influence of misspecified model components \citep{Carmona2020, Yu2022}. Intuitively, variations in generative factors that are described by the known physics should not affect the data-driven subsets of the latent variables $\bm{z}_\mathrm{c}$ and $\bm{z}_\mathrm{y}$, while variations in generative factors not included in the known physics should still be preferentially captured by the physics-grounded subset of the latent variables. The degree to which this is achieved can be evaluated by comparing the amount of information captured by each subset $\bm{z}_\mathrm{x}$, $\bm{z}_\mathrm{c}$ and $\bm{z}_\mathrm{y}$ about a specified generative factor. When a subset of the latent variables is informative about a generative factor, it should be possible to train a regressor to predict the value of the generative factor from samples drawn from this subset of the latent variables. Based on this, we propose the following procedure to assess the amount of information about a given generative factor that is encoded in a subset of the latent variables for the trained model:

\begin{itemize}
    \item [1. ] Draw two sets of samples of generative factors $\{ ( \bm{s}^{(i)}_\mathrm{x}, \bm{s}^{(i)}_\mathrm{c}, \bm{s}^{(i)}_\mathrm{y} ) \}_{i=1}^{N_\mathrm{train}}$ and $\{ (\bm{s}'^{(i)}_\mathrm{x}, \bm{s}'^{(i)}_\mathrm{c}, \bm{s}'^{(i)}_\mathrm{y})\}_{i=1}^{N_\mathrm{test}}$ from $p_{\mathrm{gt}}(\bm{s}_\mathrm{x}, \bm{s}_\mathrm{c}, \bm{s}_\mathrm{y})$ and generate two datasets $D = \{ \bm{x}_i \}_{i=1}^{N_\mathrm{train}}$ and $D' = \{ \bm{x}_i' \}_{i=1}^{N_\mathrm{test}}$ of response measurements from the ground truth generative process.
    \item[2. ] Draw a single sample from each of the approximate posterior distributions $\bm{z}_i \sim q_\phi(\bm{z}_i | \bm{x}_i)$ and $\bm{z}'_i \sim q_\phi(\bm{z}'_i | \bm{x}'_i)$ for each $\bm{x}_i \in D$ and $\bm{x}'_i \in D'$ respectively, using the trained model. This yields two sets of samples from the latent variables $\{ ( \bm{z}^{(i)}_\mathrm{x}, \bm{z}^{(i)}_\mathrm{c}, \bm{z}^{(i)}_\mathrm{y} ) \}_{i=1}^{N_\mathrm{train}}$, and $\{ ( \bm{z}'^{(i)}_\mathrm{x}, \bm{z}'^{(i)}_\mathrm{c}, \bm{z}'^{(i)}_\mathrm{y} ) \}_{i=1}^{N_\mathrm{test}}$.
    \item[3. ] Train a regressor to predict the value of each set of generative factors $\{s^{(i)}_j \}_{i=1}^{N_\mathrm{train}}$ from each subset $\{ \bm{z}^{(i)}_\mathrm{x} \}_{i=1}^{N_\mathrm{train}}$, $\{ \bm{z}^{(i)}_\mathrm{c} \}_{i=1}^{N_\mathrm{train}}$ and $\{ \bm{z}^{(i)}_\mathrm{y} \}_{i=1}^{N_\mathrm{train}}$, for $j = 1, ..., N_f$, where $N_f$ is the number of generative factors. This process yields $3 \times  N_\mathrm{f}$ regressors.

    \item[4. ] Compute the $R^2$ value between each subset $\{ \bm{z}'^{(i)}_\mathrm{x} \}_{i=1}^{N_\mathrm{test}}$, $\{ \bm{z}'^{(i)}_\mathrm{c} \}_{i=1}^{N_\mathrm{test}}$ and $\{ \bm{z}'^{(i)}_\mathrm{y} \}_{i=1}^{N_\mathrm{test}}$ and each set of generative factors $\{s'^{(i)}_j \}_{i=1}^{N_\mathrm{test}}$ using the corresponding trained regression model.
\end{itemize}

This procedure yields $N_f$ sets of pair-wise $R^2$ values $\{ R^2_{\bm{z}_\mathrm{x} \rightarrow s_j}, R^2_{\bm{z}_\mathrm{c} \rightarrow s_j}, R^2_{\bm{z}_\mathrm{y} \rightarrow s_j} \}_{j=1}^{N_f}$ with each of the $N_f$ sets corresponding to a single generative factor. A more informative subset of the latent variables should yield a more accurate regressor than an uninformative subset, and therefore also a higher $R^2$ value. It is emphasized that the metric described here is only intended to be a surrogate quantity for the amount of information encoded in each subset of the latent variables, and not a metric of disentanglement. Furthermore, this metric requires access to the ground truth distribution and data generating process, and is therefore not generally applicable.

\section{Previous work}
\label{section:previous_work}
Deep generative models such as VAE describe a mapping between a high-dimensional data manifold, and a low dimensional latent representation. Generative factors in the data are not generally controlled by individual dimensions of the latent variables, nor are they amenable to human interpretation or semantically meaningful. Disentangled representation learning is aimed at learning representations where perturbations of individual dimensions of the latent space correspond to interpretable perturbations of the data \citep{Esmaeili2019}. Approaches for disentangled representation learning can be broadly classified as either unsupervised, where disentanglement is achieved through the use of additional regularization terms on the ELBO, or supervised, semi-supervised, and weakly supervised methods that utilize additional observables or other information. For a comprehensive review of representation learning and of the different approaches, focusing on VAE, the reader is referred to \cite{Bengio2014} and \cite{Tschannen2018} respectively.

Unsupervised disentanglement necessarily relies on inductive bias and implicit supervision \citep{Locatello2019}. A common approach is to adjust the relative importance of the KLD and reconstruction error terms. In the $\beta$-VAE architecture \citep{Higgins2016}, the KLD is scaled by a factor $\beta \geq 1$ that determines how much the approximate posterior is penalized for deviating from the prior distribution, limiting the capacity of the latent distribution and encouraging the latent variables to be factorized, at the expense of reconstruction quality \citep{Burgess2018}. Other approaches involve weighting the importance of the \textit{total correlation} term \citep{Watanabe1960}, a component of the KLD that quantifies and penalizes dependence between the dimensions of the aggregated posterior distribution (i.e. the posterior distribution marginalized over the entire dataset). These approaches avoid the high computational cost associated with estimating the total correlation by utilizing stochastic approximations based on mini-batches \citep{Chen2018} and adversarial density-ratio estimation \citep{Kim2018}. The \textit{InfoVAE} approach proposed by \cite{Zhao2018} involves scaling specific terms in the ELBO, coupled with an additional term that promotes maximization of the MI between the inputs and latent variables. Other approaches for unsupervised disentanglement have been proposed in the literature, such as enforcing independence between and within groups of latent variables \citep{Esmaeili2019}, extending the standard architecture with adversarial components \citep{Larsen2016} and additional decoders \citep{Ding2020} and using sparsity inducing priors \citep{Tonolini2020}. Several works utilize additional observables in a fully supervised \citep{Hadad2018, Sun2022, Debbagh2023}, or semi-supervised \citep{Louizos2017} manner, combined with inductive biases in the form of structured models \citep{Siddharth2017} and penalties on dependence \citep{Lopez2018} to promote disentanglement or invariance to nuisance factors. It has also been shown \citep{Achille2017} that disentanglement is closely related to the information bottleneck theory introduced by \cite{Tishby2000}, and later extended to the variational setting by \cite{Alemi2019}. Finally, the more general notion of decomposition, that admits disentanglement as a special case, was introduced by \cite{Mathieu2019}.

It has been hypothesized that representation learning approaches can be particularly useful in domain adaptation and transfer learning tasks due to their ability to capture underlying generative factors in data that are shared between tasks \citep{Bengio2014}. The gradient reversal approach utilized in this work was originally proposed to tackle domain adaptation for image classification \citep{Ganin2015, Ganin2016}. Similar approaches have been extended to the setting of multi-view and multi-modal learning \citep{Federici2020, Mondal2023, Aguerri2019, Hwang2020}. Recent work has also explored the application of adversarial domain adaptation techniques to structural damage identification \citep{Wang2022}. Finally, it is important to note that our proposed architecture is similar to the Domain Invariant Variational Autoencoder (DIVA) proposed by \cite{Ilse2020}, from which we adopt part of our terminology. DIVA is targeted towards invariant representation learning and domain generalization in a purely data-driven setting and does not utilize adversarial training, instead explicitly imposing an independence assumption between subsets of the latent variables.

VI offers a balance between accuracy and computational tractability. Combined with the inherent regularization of the Bayesian framework, these properties are particularly advantageous in the modeling of physical systems \citep{GlynDavies2025}. As a result, incorporating physical knowledge in VAE has received significant attention, and various physics-informed formulations have been proposed depending on the task of interest. Notably, \cite{Walker2024} present an approach for utilizing known physics to discover shared information in multi-modal data. The UQ-VAE \citep{Goh2022} combines known governing equations and prior distributions over parameters of interest with paired input-output measurements to achieve computationally efficient uncertainty quantification for systems described by partial differential equations. Formulations of VAE that take advantage of known governing equations have also been proposed for solving forward and inverse problems in stochastic differential equations \citep{Zhong2023, Shin2023}. In the context of surrogate modeling, \cite{Rixner2021} introduce the notion of virtual observables as a means of encoding physical knowledge into probabilistic generative models. Despite these advances, the issue of balancing physics-based and data-driven components in VAE has received relatively limited attention. This issue is addressed by \cite{Takeishi2021} for systems described by ordinary differential equations. A similar setting is investigated in \cite{Linial2021} and \cite{Yildiz2019}, with the latter introducing a regularized objective to ensure consistency of the latent space with the known physics.

\section{Synthetic case studies}
\label{section:case_studies}

Three synthetic case studies of different complexity, illustrated in \autoref{fig:examples_overview}, are discussed in detail throughout this section. The case study objectives, the definition of the physics-based models, the procedure used to generate the synthetic data, and details of the model implementation and visualization are provided below. For the purposes of reproducibility, the code needed to replicate the examples is made available on GitHub\footnote{\url{https://github.com/JanKoune/DPI-VAE}}. Additional information regarding the architecture, variable transformations, data, optimization, and visualization is provided in Appendix \ref{appendix:appendix_A}.

\paragraph*{Case study objectives}
Each case study addresses a different set of challenges. The beam case study demonstrates that the proposed approach preferentially utilizes the known physics, yielding an interpretable and parsimonious representation of the physical system. The oscillator case study highlights how the adversarial training can prevent the model from learning arbitrary components of the response in a data-driven manner, and investigates the impact of the GRL hyperparameter. Finally, the bridge case study demonstrates the feasibility of using the model for damage detection in a more complex synthetic case, and compares the performance to that of existing data-driven approaches. It is noted that the case studies are only meant as didactic examples, intended to elaborate the issues with combining physics-based and data-driven components in VAE, provide intuition about the interaction between these components, and demonstrate the behavior of the model. Therefore, emphasis is placed on clarity rather than realism.

\paragraph*{Physics-based models}
Three separate physics-based models are considered for every case study: A high-fidelity \textit{simulator}, a \textit{full} model, and a \textit{nominal} model. The simulator is an accurate but generally computationally expensive model of the physical system, typically in the form of a Finite Element (FE) model, used to train the full and nominal models for each case study. The full model is a computationally efficient surrogate model of the ground truth data generating process: one or more structures with varying physical characteristics, subject to operational and environmental conditions, damage and degradation. To produce the training dataset for the full model, the simulator is evaluated on a set of generative factors $\{( \bm{s}^{(i)}_{\mathrm{x}}, \bm{s}^{(i)}_{\mathrm{c}}, \bm{s}^{(i)}_{\mathrm{y}} )\}_{i=1}^{N_\mathrm{full}}$, sampled uniformly and independently from prescribed ranges of values. The ranges are chosen to provide sufficient coverage over the support of the corresponding ground truth distribution $p_{\mathrm{gt}}(\bm{s}_{\mathrm{x}}, \bm{s}_{\mathrm{c}}, \bm{s}_{\mathrm{y}})$. The full model is then obtained by fitting a NN-based surrogate to the dataset composed of $N_\mathrm{full}$ input-output pairs $D_{\mathrm{full}} = \{( \bm{s}^{(i)}_{\mathrm{x}}, \bm{s}^{(i)}_{\mathrm{c}}, \bm{s}^{(i)}_{\mathrm{y}}, \bm{x}^{(i)} )\}_{i=1}^{N_{\mathrm{full}}}$ obtained from the simulator. Using a NN as the forward model for the data generating process enables the efficient visualization of the latent space and the reconstructions generated by the VAE for different inputs, simplifies the generation of test data to evaluate the performance of the VAE, and makes it possible to account for randomness in the hyperparameter initialization and data generation by averaging results over multiple runs with i.i.d. datasets. The nominal model corresponds to the available incomplete representation of the physics of the system under investigation. When an analytical expression describing the partially known physics is available, this is used as the nominal model. Alternatively, the nominal model is built by training a NN-based surrogate on a limited dataset $D_{\mathrm{nom}} = \{ (\bm{s}^{(i)}_{\mathrm{x}}, \bm{x}^{(i)} ) \}_{i=1}^{N_{\mathrm{nom}}}$, obtained by evaluating the simulator only on the physics-based subset of the generative factors $\{ \bm{s}^{(i)}_{\mathrm{x}} \}_{i=1}^{N_\mathrm{nom}}$, while $\bm{s}_{\mathrm{c}}$ and $\bm{s}_{\mathrm{y}}$ are set to a constant reference value corresponding to the nominal condition of the structure. 

\begin{figure}[htb!]
	\centering
    \FIG{\includegraphics[width=0.90\textwidth]{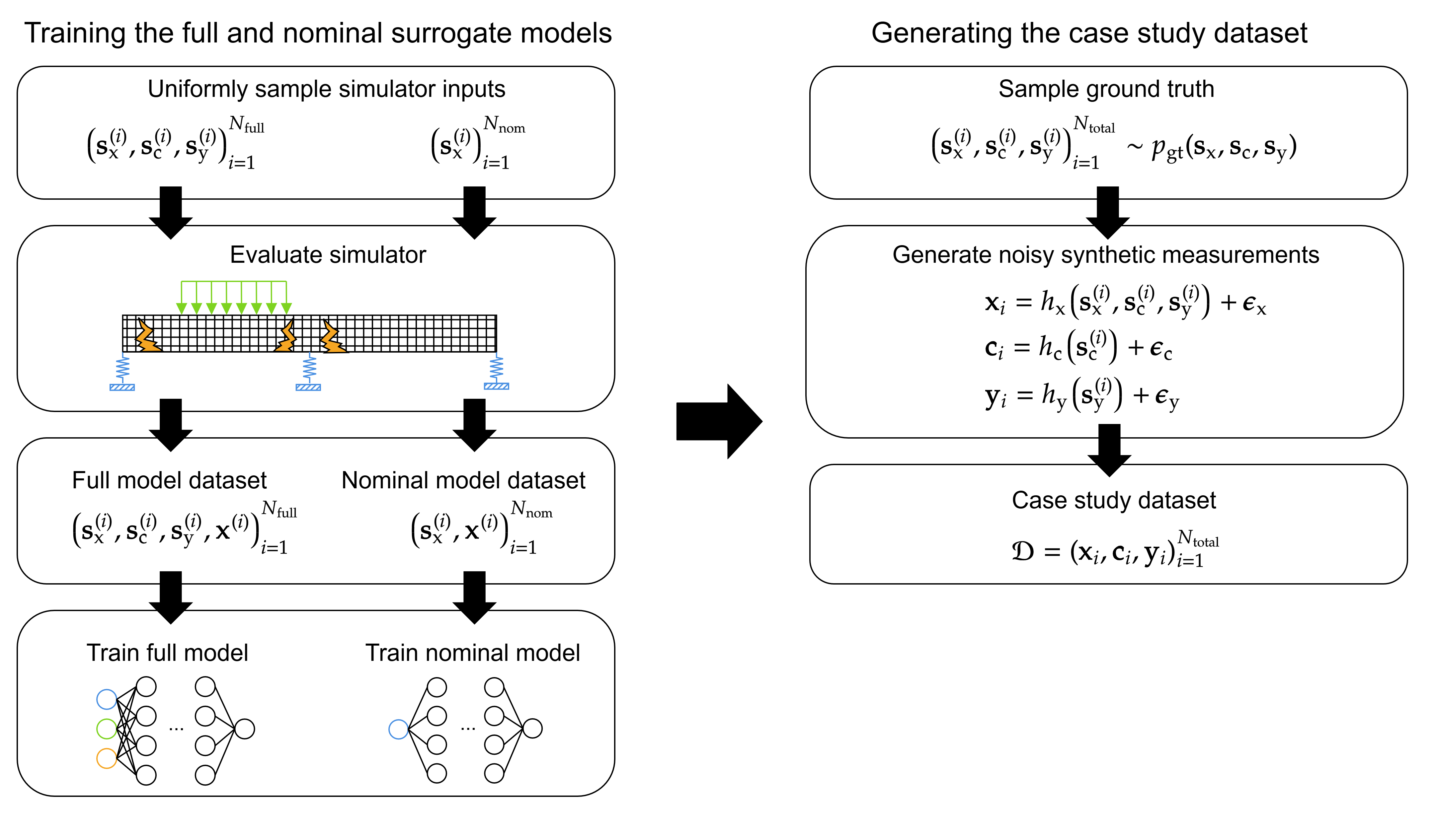}}
	\caption{Illustration of the procedure used to obtain the full and nominal physics-based models (left), and to generate the datasets used in the case studies (right).}
	\label{fig:case_definition}
\end{figure}

\paragraph*{Synthetic data generation}
The VAE is trained and validated on a dataset composed of $N_\mathrm{total} = N_\mathrm{train} + N_{\mathrm{val}}$ triplets of observables $\mathcal{D} = \{ (\bm{x}_i, \bm{c_i}, \bm{y_i} ) \}_{i=1}^{N_\mathrm{total}}$. This dataset is generated by first drawing samples of the generative factors from the ground truth distribution $( \bm{s}^{(i)}_{\mathrm{x}}, \bm{s}^{(i)}_{\mathrm{c}}, \bm{s}^{(i)}_{\mathrm{y}} ) \sim{p_\mathrm{gt}(\bm{s}_{\mathrm{x}}, \bm{s}_{\mathrm{c}}, \bm{s}_{\mathrm{y}})}$ for $i=1, ...,N_\mathrm{total}$, applying a set of deterministic transformations, and subsequently adding i.i.d. samples of zero-mean Gaussian white noise $\bm{\epsilon}_\mathrm{x} \sim \mathcal{N}(\bm{0}, \bm{\sigma}^2_\mathrm{x} \bm{I})$, $\bm{\epsilon}_\mathrm{c} \sim \mathcal{N}(\bm{0}, \bm{\sigma}^2_\mathrm{c} \bm{I})$ and $\bm{\epsilon}_\mathrm{y} \sim \mathcal{N}(\bm{0}, \bm{\sigma}^2_\mathrm{y} \bm{I})$. Denoting the full model as $h_\mathrm{x}(\cdot)$, the response observables are obtained as $\bm{x}_i = h_{\mathrm{x}}(\bm{s}^{(i)}_\mathrm{x}, \bm{s}^{(i)}_\mathrm{c}, \bm{s}^{(i)}_\mathrm{y}) + \bm{\epsilon}_{\mathrm{x}}$. The domain and class observables are obtained as $\bm{c}_i = h_{\mathrm{c}}(\bm{s}^{(i)}_\mathrm{c}) + \bm{\epsilon}_\mathrm{c}$ and $\bm{y}_i = h_{\mathrm{y}}(\bm{s}^{(i)}_\mathrm{y}) + \bm{\epsilon}_\mathrm{y}$ respectively. This procedure is illustrated in \autoref{fig:case_definition}. 

\paragraph*{Implementation details}
For all the case studies presented in this section, $h_\mathrm{c}$ and $h_{\mathrm{y}}$ are taken as the identity function for simplicity. Furthermore we use $N_\mathrm{train} = 1024$ and $N_\mathrm{val} = 512$, and consider no other regularization except for the GRL, i.e. $\beta = \alpha_\mathrm{x} = \alpha_\mathrm{c} = \alpha_\mathrm{y} = 1.0$. Unless stated otherwise, the number of the domain and class latent variables are taken to be twice the number of domain and class generative factors\footnote{The intention behind this choice is to avoid biasing the model towards a disentangled representation by matching the number of latent variables to the ground truth generative factors, ensuring that any disentanglement in the learned representation is not a consequence of limited latent space capacity.} To enable the formulation of problems with bounded latent variables and to ensure a stable optimization procedure, the physics-grounded latent variables are obtained through a sequence of invertible transformations applied to the encoder output, mapping samples from an unbounded base latent space to the target latent space. All physics-grounded latent variables are constrained to lie within ranges that ensure consistency with the underlying physics of each system.

\paragraph*{Visualization}
A particularly useful tool for assessing the learned representation is to ``traverse'' the latent space and the space of reconstructions of the VAE \citep{Kim2018, Higgins2016}. This can be achieved by generating synthetic data while interpolating over a specified generative factor, and setting the remaining generative factors to a constant reference value. The VAE is then evaluated on the generated data, yielding samples from the latent space, realizations of the reconstructed input $\hat{\bm{x}}$, and the mean physics-based and data-driven components $\hat{\bm{x}}_\mathrm{p}$ and $\hat{\bm{x}}_\mathrm{d}$. Each generative factor is linearly interpolated within the $1^\mathrm{st}$ and $99^\mathrm{th}$ percentiles of the corresponding ground truth distribution.

\subsection{Beam case study}
\label{section:beam_example}

\subsubsection{Case study description}
The case study consists of a beam with fixed length $L=1.0$ m and a point load with magnitude $F=1.0$ N acting on an uncertain position $x_\mathrm{F}$ along the length of the beam. The material is linear elastic with uncertain Young's modulus $E$, Poisson ratio $\nu=0.3$, area moment of inertia $I = 2\cdot10^{-6}$ m$^4$ and cross-sectional area $A = 2.4 \cdot 10^{-3}$ m$^2$. The rotational stiffness of the right-hand side support is temperature dependent, with the dependence modeled as an increase in the rotational stiffness of the support at lower temperatures. The relationship between the temperature and the support rotational stiffness is formulated as $\log k_\mathrm{r} = 8- \frac{10}{1 +  e^{-T/2}}$. The beam is subject to variability in the vertical stiffness of the right-hand side support, e.g. due to damage or a deficiency of the support, simulated as a translational spring boundary condition with stiffness $k_\mathrm{v}$. This quantity can span several orders of magnitude, and therefore we parametrize the model using $\log k_\mathrm{v}$ instead. The beam is equipped with $d_\mathrm{x} = 32$ sensors measuring the vertical displacement, equally spaced along the length as shown in \autoref{fig:examples_overview}(a).

The Young’s modulus $E$ and the position of the point load $x_\mathrm{F}$ are considered as uncertain latent variables, such that $\bm{z}_\mathrm{x} = (E, x_\mathrm{F})$. It is assumed that the temperature is an observed domain variable such that $\bm{c} = (T)$, and that the vertical spring log-stiffness $\log k_\mathrm{v}$ is taken as a class variable representing damage in the structure, such that $\bm{y} = (\log k_\mathrm{v})$. Since the class variable $\bm{y}$ represents damage in the structure it will not be quantitatively measurable. In a realistic scenario, observations of the condition of the support on a qualitative scale (e.g. from $0$ representing no damage to $5$ denoting a fully damaged support) might be available. In this example we simplistically consider $\bm{y}$ as the ground truth value of $\log k_\mathrm{v}$ with some added noise. The variable symbols, units, types, as well as the prior distributions over the physics-grounded latent variables and the ground truth distributions of the generative factors used to generate the training data are summarized in \autoref{tab:beam_variables}. We additionally provide a reference value which is used to produce the figures as discussed in Section \ref{section:case_studies}. To ensure physical consistency and to avoid numerical issues, the Young's modulus is truncated below a small positive value, and the load position $x_\mathrm{F}$ is restricted to the range $(0, 1)$.

\begin{table}[htb!]
\centering
\caption{Summary of generative factors and the corresponding ground truth and prior distributions.}
\label{tab:beam_variables}
\begin{tabular}{@{}cccccc@{}}
\toprule
Variable & Unit & Type & Prior distribution & Ground truth & Reference value  \\ 
\midrule
$E$ & MPa & Physical & $\mathcal{N}(4.0, 1.0)$ & $\mathcal{U}(2.5, 4.5)$ & 3.0 \\
$x_\mathrm{F}$ & m & Physical & $\mathcal{N}(0.5, 0.04)$ & $\mathcal{U}(0.3, 0.7)$ & 0.5 \\
$\log k_\mathrm{v}$ & N/m & Class & - & $\mathcal{U}(6.0, 8.0)$ & 8.0 \\
$T$ & C$^o$ & Domain & - & $\mathcal{U}(-11.0, 5.0)$ & 5.0 \\
\bottomrule
\end{tabular}
\end{table}

A partial description of the physics is available, in the form of an analytical expression for the vertical deflection of a simply supported Euler-Bernoulli beam with a point load acting at $x_\mathrm{F}$:

\begin{equation}
    \label{eq:euler_bernoulli_point_load}
    w(x) = 
        \begin{cases}
            \frac{P b x (L^2 - b^2 - x^2)}{6 L E I}, & 0 \leq x \leq x_\mathrm{F} \\
            \frac{P b x (L^2 - b^2 - x^2)}{6 L E I} + \frac{P(x - x_\mathrm{F})^3}{6 E I}, & x_\mathrm{F} < x \leq L
        \end{cases}
\end{equation}

\noindent
where $b = L - x_\mathrm{F}$, and the non-bold $x$ refers to the position along the beam. This nominal model represents the beam in the undamaged condition at a reference temperature, and is directly incorporated in the physics-based branch of the VAE decoder.

Following the procedure described in Section \ref{section:case_studies}, the full model (trained on input-output pairs from an FE-based simulator) is used to produce synthetic data by first drawing samples of the input parameters from the ground truth distribution, and subsequently contaminating the resulting model predictions with zero-mean Gaussian white noise with standard deviation $\sigma_\mathrm{x} = 0.02$ m. The dataset used to train the VAE is composed of $N_\mathrm{train}$ measurements of the beam displacement $\bm{x} = \{\bm{x}_i\}_{i=1}^{N_\mathrm{train}}$ where each element $\bm{x}_i$ is a vector of length $d_\mathrm{x} = 32$. The domain and class observables $\bm{c} = \{\bm{c}_i\}_{i=1}^{N_\mathrm{train}}$ and $\bm{y} = \{\bm{y}_i\}_{i=1}^{N_\mathrm{train}}$ are obtained as the ground truth values used to generate the dataset, with the addition of i.i.d. samples of Gaussian white noise with standard deviations of $\sigma_c = \sigma_y = 0.02$ respectively. The dimensionality of the domain and class latent variables is taken as $d_{z_\mathrm{c}} = d_{z_\mathrm{y}} = 2$.

\subsubsection{Qualitative assessment of disentanglement}
\begin{figure}[htb!]
    \centering
    \FIG{\includegraphics[width=1.0\textwidth]{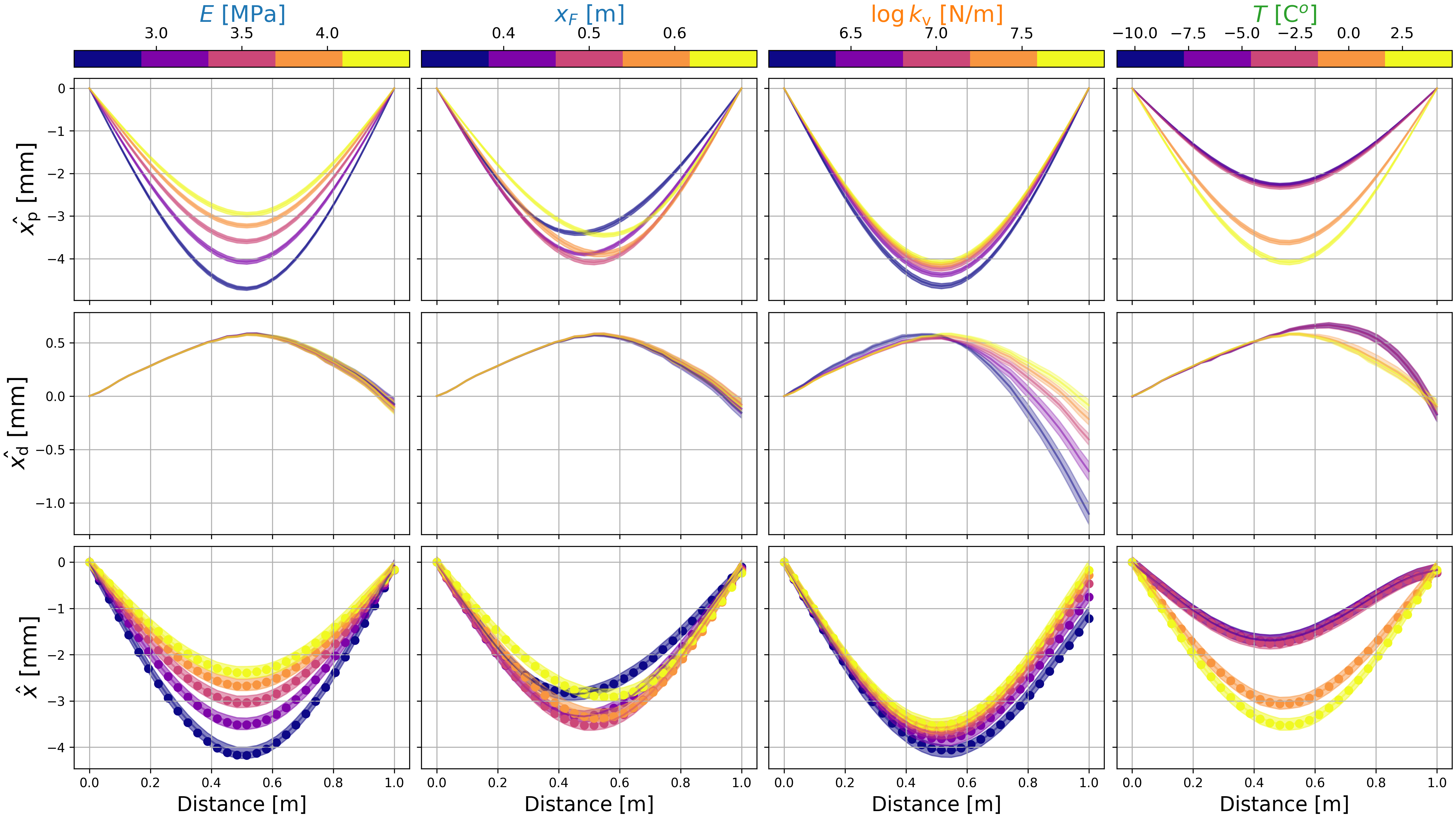}}
    \caption{Mean prediction and $\pm 2\sigma$ uncertainty bounds for the physics-based $\hat{\bm{x}}_\mathrm{p}$ and data-driven $\hat{\bm{x}}_\mathrm{d}$ components, and combined prediction $\hat{\bm{x}}$ while traversing the generative factors. The input response measurements are denoted as dots in the bottom row.}
    \label{fig:beam_pred_1}
\end{figure}

After training, the disentanglement between physics-grounded and data-driven components is qualitatively assessed by examining the latent space and samples of the reconstructed response of the beam. The predicted physics-based $\hat{\bm{x}}_{\mathrm{p}}$ and data-driven $\hat{\bm{x}}_{\mathrm{d}}$ components, as well as the combined prediction $\hat{\bm{x}}$ are shown in \autoref{fig:beam_pred_1}. It can be observed that the data-driven component of the reconstruction $\hat{\bm{x}}_{\mathrm{d}}$ (middle row) is invariant to changes in $E$ and $x_F$ contributing only a constant deformed shape to the total predicted response. On the other hand, the physics-based model captures variability in both the physics-grounded generative factors $\bm{s}_\mathrm{x}=( E, x_\mathrm{F} )$, but also domain and class generative factors $\bm{s}_{\mathrm{c}} = ( T )$ and $\bm{s}_{\mathrm{y}} = ( \log k_{\mathrm{v}} )$. In contrast to the behavior of the unconstrained VAE, presented in Section \ref{section:challenges}, here the model preferentially utilizes the known physics. Only the variability in the measured response due to $\log{k_\mathrm{v}}$ and $T$ that can not be captured by the physics-based model is accounted for by the data-driven part of the decoder, indicating that the model can disentangle components of the response that can be attributed to the known physics from those that cannot. A key aspect of the adversarial training is the degree to which it allows interaction between the physics-based and data-driven components of the prediction. In this case study, the additional displacement of the beam due to the reduced vertical stiffness of the right-hand side support will also depend on the load position $x_\mathrm{F}$. More positive values of $\lambda$ tend to prevent the model from capturing this interaction, whereas more negative values enable it but may result in the data-driven components overriding the known physics.

\begin{figure}[htb!]
	\centering
    \begin{subfigure}{0.45\textwidth}
    	\includegraphics[width=1.0\textwidth]{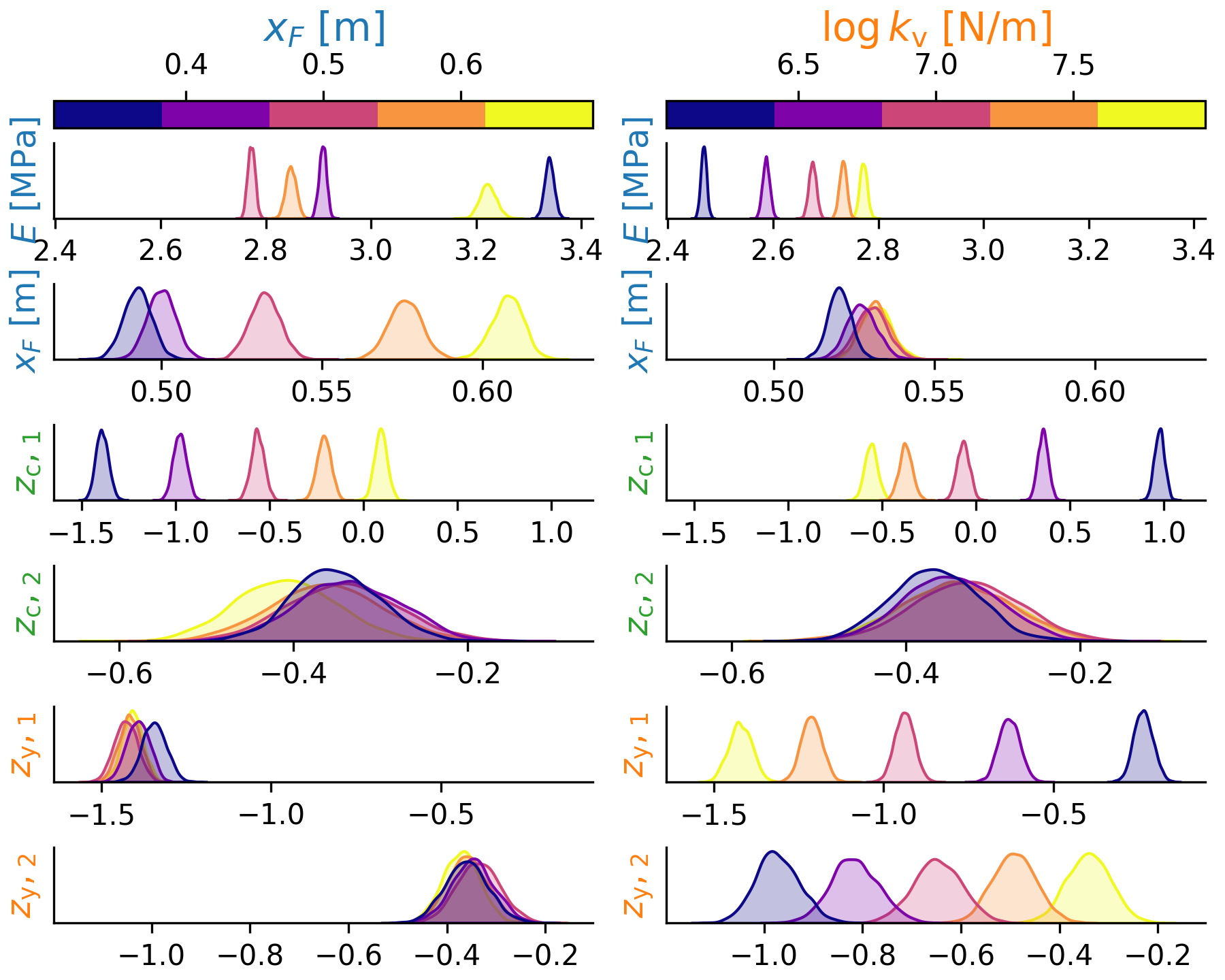}
        \caption{Marginal approximate posterior without adversarial training ($\lambda = -1$).}
    	\label{fig:beam_latent_2a}
    \end{subfigure}
    \begin{subfigure}{0.45\textwidth}
    	\includegraphics[width=1.0\textwidth]{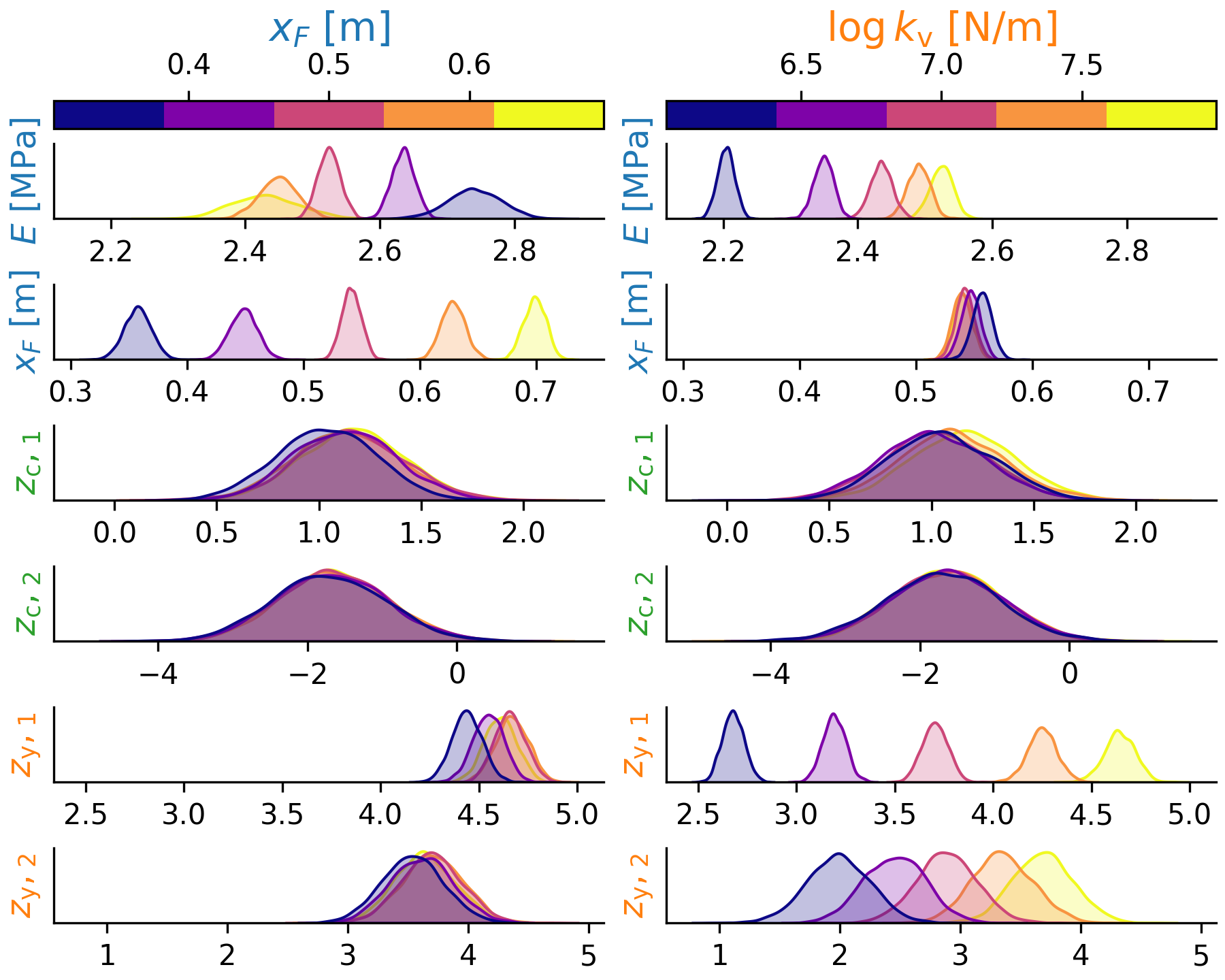}
        \caption{Marginal approximate posterior with adversarial training ($\lambda = 1/256$).}
    	\label{fig:beam_latent_2b}
    \end{subfigure}
\caption{Visualizations of the VAE latent space during traversal of the generative factors $x_\mathrm{F}$ and $\log k_{\mathrm{v}}$. Each column corresponds to variation of a single generative factor, and each row shows the marginal approximate posterior distribution of a single latent variable.}
\label{fig:beam_latent_2}
\end{figure}

To further highlight the impact of the GRL, the latent space traversals of the unconstrained model and the model trained adversarially are compared in \autoref{fig:beam_latent_2}. Without adversarial training, the domain latent variables $\bm{z}_{\mathrm{c}}$ encode the variability in the load position  $x_\mathrm{F}$ as shown in \autoref{fig:beam_latent_2}(a), providing the data-driven decoder components with the information needed to reconstruct this component of the measured response and resulting in an entangled representation, as discussed in Section \ref{section:challenges}. In contrast, when $\lambda = 1/256$ the adversarial training results in a posterior distribution over $\bm{z}_{\mathrm{c}}$ that is invariant to changes in $x_\mathrm{F}$. Instead, the variability is captured by the corresponding physics-grounded latent variable, as shown in \autoref{fig:beam_latent_2}(b), indicating disentanglement of the physics-grounded and domain generative factors. The results shown previously suggest that the latent bottleneck architecture and GRL regularization result in a sparse and parsimonious representation of the physical system, and can yield domain and class latent variables that are invariant to changes in the underlying physics. The influence of the GRL hyperparameter $\lambda$ is further investigated in the oscillator case study presented below.

\subsection{Oscillator case study}
\label{section:oscillator_example}

\subsubsection{Case study description}
This example demonstrates how the adversarial training prevents the model from compensating for all discrepancies between the physics-based model predictions and measurements. Suppose that a mass-spring-dashpot system undergoes damped harmonic motion, starting from an initial displaced position $x_0$, with no external excitation. It is assumed that each experiment is performed under varying temperature $T$, which is taken as the domain variable. The temperature affects the spring stiffness through the relationship $k(T) =  k_{\mathrm{ref}} + \alpha_\mathrm{T} (T_\mathrm{ref} - T)$, with $T_\mathrm{ref} = 20.0$ C$^o$ and $\alpha_\mathrm{T} = 0.01$. The reference spring stiffness at $T_\mathrm{ref} = 20.0$ C$^o$ is assumed known and equal to $k_{\mathrm{ref}} = 1.0$ N/m. The mass $m$ is considered unknown and treated as a physics-grounded latent variable to be inferred from data. The viscous damping coefficient $c_\mathrm{d}$ is taken to define the class of the system. Finally, it is assumed that the observations are subject to an unknown confounding influence in the form of small random perturbations of the initial displacement $x_0$. A summary of the generative factors, the prior distribution and the ground truth distribution is provided in \autoref{tab:oscillator_variables}.

\begin{table}[htb!]
\centering
\caption{Summary of generative factors and the corresponding ground truth and prior distributions.}
\label{tab:oscillator_variables}
\begin{tabular}{@{}cccccc@{}}
\toprule
Variable & Unit & Type & Prior distribution & Ground truth & Reference value \\ 
\midrule
$m$ & kg & Physical & $\mathcal{U}(1.0, 2.0)$ & $\mathcal{U}(1.2, 1.8)$ & 1.5 \\
$c_\mathrm{d}$ & kg / s & Class & - & $\mathcal{U}(0.0, 2.0)$ & 0.0 \\
$T$ & C$^o$ & Domain & - & $\mathcal{U}(0.0, 40.0)$ & 20.0 \\
$x_0$ & m & Unknown & - & $\mathcal{U}(0.9, 1.1)$ & 1.0 \\
\bottomrule
\end{tabular}
\end{table}

The equation of motion describing the system can be written as:

\begin{equation}
    \label{eq:pendulum_eom}
    m \frac{\mathrm{d}^2 x(t)}{\mathrm{d}t^2} + c_\mathrm{d} \frac{\mathrm{d}x (t)}{\mathrm{d}t} + k(T) x (t) = 0
\end{equation}

\noindent
A partial description of the physics is available in the form of an analytical solution under the assumption that the initial displacement is $x_0 = 1.0$ m, and the initial velocity is $\dot{x}_0 = 0.0$ m/s for all experiments, and that there is no damping affecting the motion of the oscillator. Furthermore, it is assumed that the relationship between temperature and stiffness is not known, and the temperature effect is therefore not included in the nominal physics-based model. Under the assumptions described previously, the displacement of the oscillator at time $t$ can be expressed as:

\begin{equation}
    \label{eq:pendulum_part_model}
    x(t) = \cos \left( \sqrt{\frac{k_\mathrm{ref}}{m}} t \right)
\end{equation}

Each triplet of observations is composed of a noisy displacement time series, and noisy measurements of the viscous damping coefficient $c_\mathrm{d}$ and temperature $T$, which are considered as class and domain variables respectively such that $\bm{c}= (T)$ and $\bm{y} = (c_\mathrm{d})$. Training and validation datasets are generated by drawing samples from the ground truth distribution and generating the oscillator displacement time-series using the full model, trained on input-output pairs simulated using the equation of motion shown in \autoref{eq:pendulum_eom}. Each of the measured time-series is a vector of $64$ measurements, equally spaced within a time interval $t \in [0, 10]$ s. The synthetic response measurements are subsequently contaminated with i.i.d. realizations of Gaussian white noise with standard deviation $\sigma_\mathrm{x} = 0.01$ m. The standard deviation of the measurement uncertainty of the domain and class observables are taken as $\sigma_\mathrm{c} = 0.01$ and $\sigma_\mathrm{y} = 0.01$ respectively. To ensure that the model has sufficient capacity to learn the unknown confounding influence if the adversarial training were not present, the dimensionality of the latent space is specified to be significantly larger than the number of ground truth generative factors. The domain and class latent space dimensions are taken as $d_{z_\mathrm{c}} = d_{z_\mathrm{y}} = 4$.

\subsubsection{Model behavior in the presence of unknown confounders}
\begin{figure}[htb!]
	\centering
    \begin{subfigure}{1.0\textwidth}
        \centering
    	\includegraphics[width=0.9\textwidth]{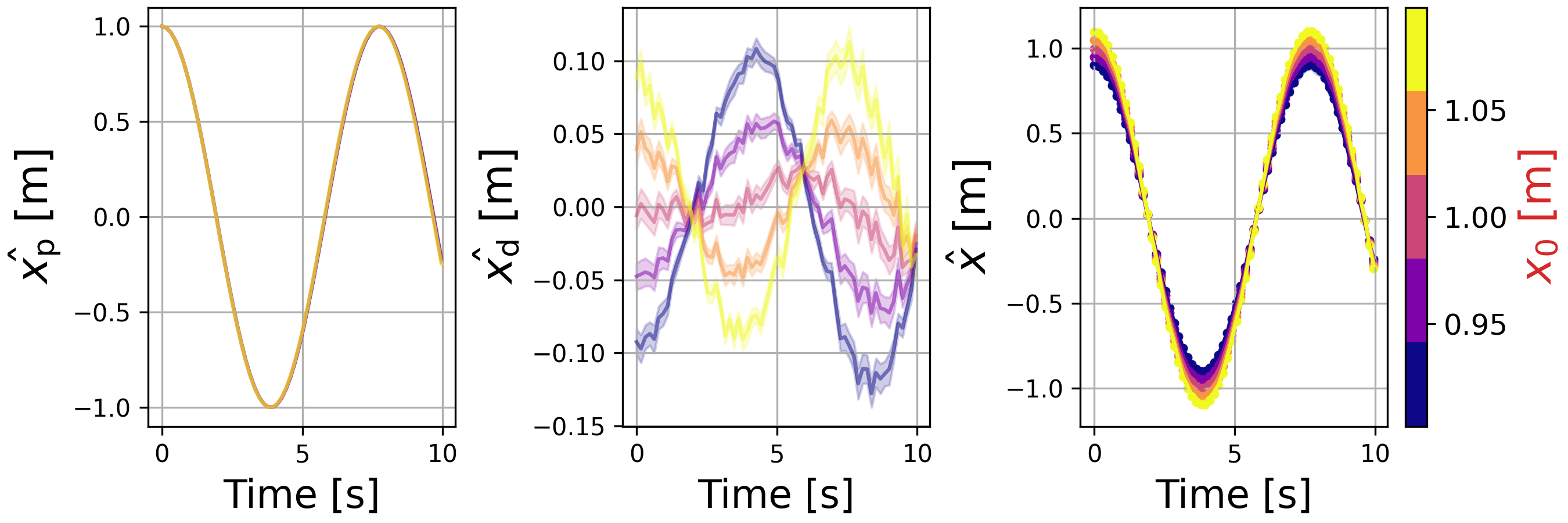}
        \caption{Mean and $\pm 2 \sigma$ uncertainty bounds of the reconstructed response without adversarial training ($\lambda = -1$).}
    	\label{fig:oscillator_pred_1a}
    \end{subfigure}
    \begin{subfigure}{1.0\textwidth}
        \centering
    	\includegraphics[width=0.9\textwidth]{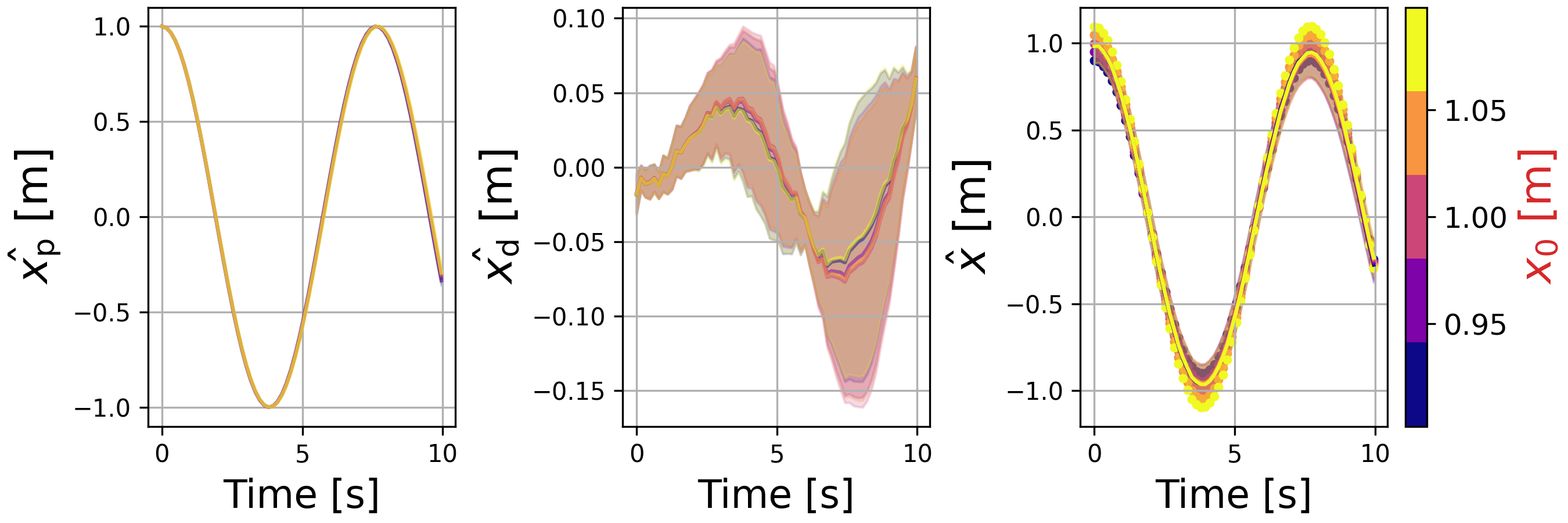}
        \caption{Mean and $\pm 2 \sigma$ uncertainty bounds of the reconstructed response with adversarial training ($\lambda = 1/128$).}
    	\label{fig:oscillator_pred_1b}
    \end{subfigure}
    \caption{Physics-based model prediction $\hat{\bm{x}}_\mathrm{p}$, data-driven model prediction $\hat{\bm{x}}_\mathrm{d}$, and combined prediction $\hat{\bm{x}}$ for varying initial displacement $x_0$. With $\lambda=-1.0$ (top) the data-driven components in the VAE are free to account for the variability in the initial position. For $\lambda=1/128$ (bottom) the model does not learn this component of the response.}
    \label{fig:oscillator_pred_1}
\end{figure}

The proposed model with no adversarial training ($\lambda = -1$) is trained and evaluated on the synthetic example. The reconstructed response obtained from a traversal of the initial displacement $x_0$, shown in \autoref{fig:oscillator_pred_1a}, highlights another issue that occurs when combining physics-based and data-driven components in VAE: Although the variability in $x_0$ can not be accounted for by the physics-based model, and there is no information in the domain or class variables regarding the value of $x_0$, the lack of regularization results in a model that is free to capture the components of the measured displacement stemming from the variability in the initial displacement $x_0$. Although in this case the effect is benign, in more complex physical systems it can result in the model learning unknown confounding influences in a non-interpretable black-box manner. When the GRL regularization is utilized (\autoref{fig:oscillator_pred_1b}), the data-driven encoder is unable to capture the variability in $x_0$, depriving the data-driven decoder from the information needed to reconstruct this component of the input measurements.

\begin{figure}[htb!]
	\centering
    \begin{subfigure}{1.0\textwidth}
        \centering
    	\includegraphics[width=0.9\textwidth]{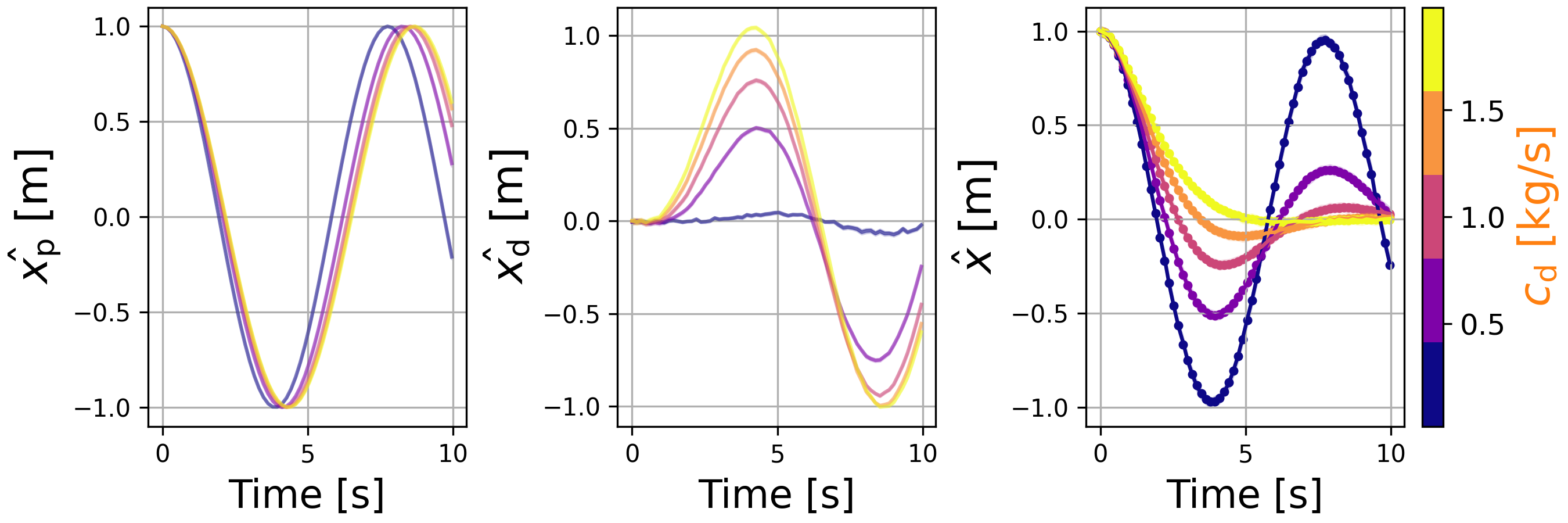}
        \caption{Mean and $\pm 2 \sigma$ uncertainty bounds of the reconstructed response without adversarial training ($\lambda = -1$).}
    	\label{fig:oscillator_pred_2a}
    \end{subfigure}
    \begin{subfigure}{1.0\textwidth}
        \centering
    	\includegraphics[width=0.9\textwidth]{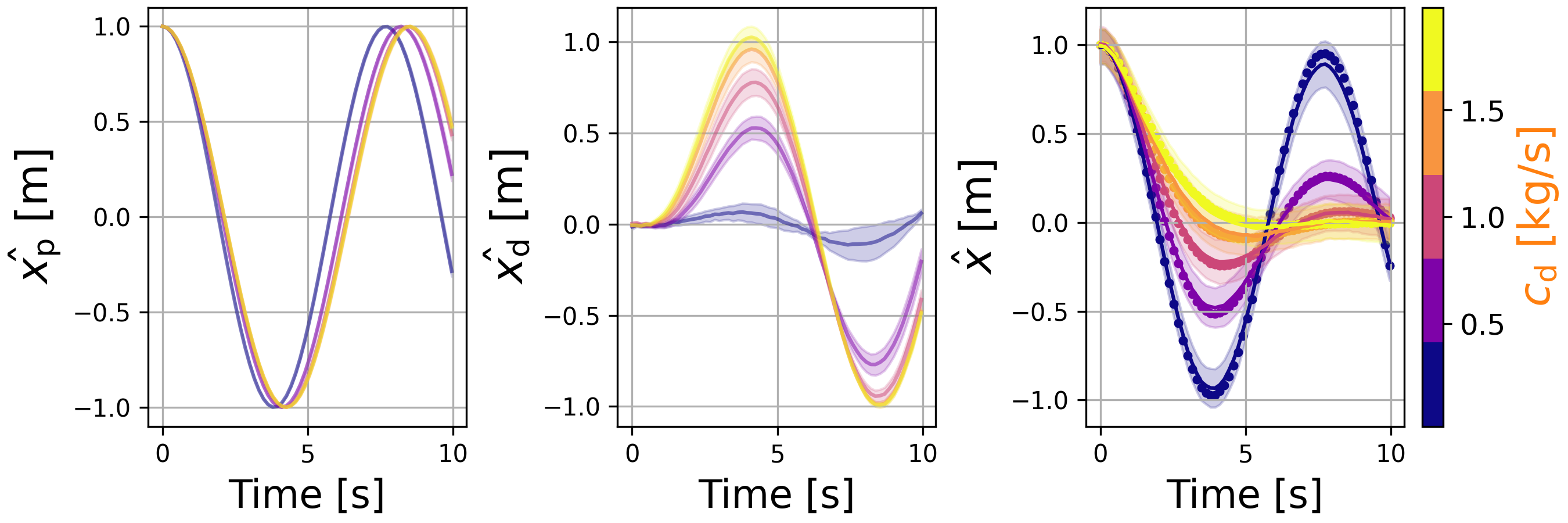}
        \caption{Mean and $\pm 2 \sigma$ uncertainty bounds of the reconstructed response with adversarial training ($\lambda = 1/128$).}
        \label{fig:oscillator_pred_2b}
    \end{subfigure}
\caption{Physics-based model prediction $\hat{\bm{x}}_\mathrm{p}$, data-driven model prediction $\hat{\bm{x}}_\mathrm{d}$, and combined prediction $\hat{\bm{x}}$ for varying viscous damping coefficient $c_\mathrm{d}$. The data-driven decoder components are prevented from fully accounting for the discrepancies between the physics-based model and measurements, resulting in wider uncertainty bounds for the proposed model.}
\label{fig:oscillator_pred_2}
\end{figure}

The results shown in \autoref{fig:oscillator_pred_1} demonstrate how the unconstrained VAE will compensate for discrepancies between the physics-based model prediction and the measurements caused by unknown confounding influences. The reason why this can be detrimental for the learning task is illustrated in \autoref{fig:oscillator_pred_2a}. The unconstrained VAE accounts for the learned confounding influence of $x_0$, which can lead to underestimation of the uncertainty over the latent variables and predictions. In contrast, when the model is trained with the adversarial objective, the encoder is prevented from learning a representation of $x_0$. The uncertainty stemming from the partial knowledge of the physics, including the unknown influence of the viscous damping and the variability in $x_0$, is more accurately accounted for in the reconstructed response, as shown in (\autoref{fig:oscillator_pred_2b}). The additional uncertainty can also be attributed to the fact that the mass-spring-dashpot system does not satisfy the additivity assumption described in Section \ref{section:challenges}. 

\subsubsection{Quantitative assessment of disentanglement}
\begin{figure}[htb!]
	\centering
	\FIG{\includegraphics[width=0.75\textwidth]{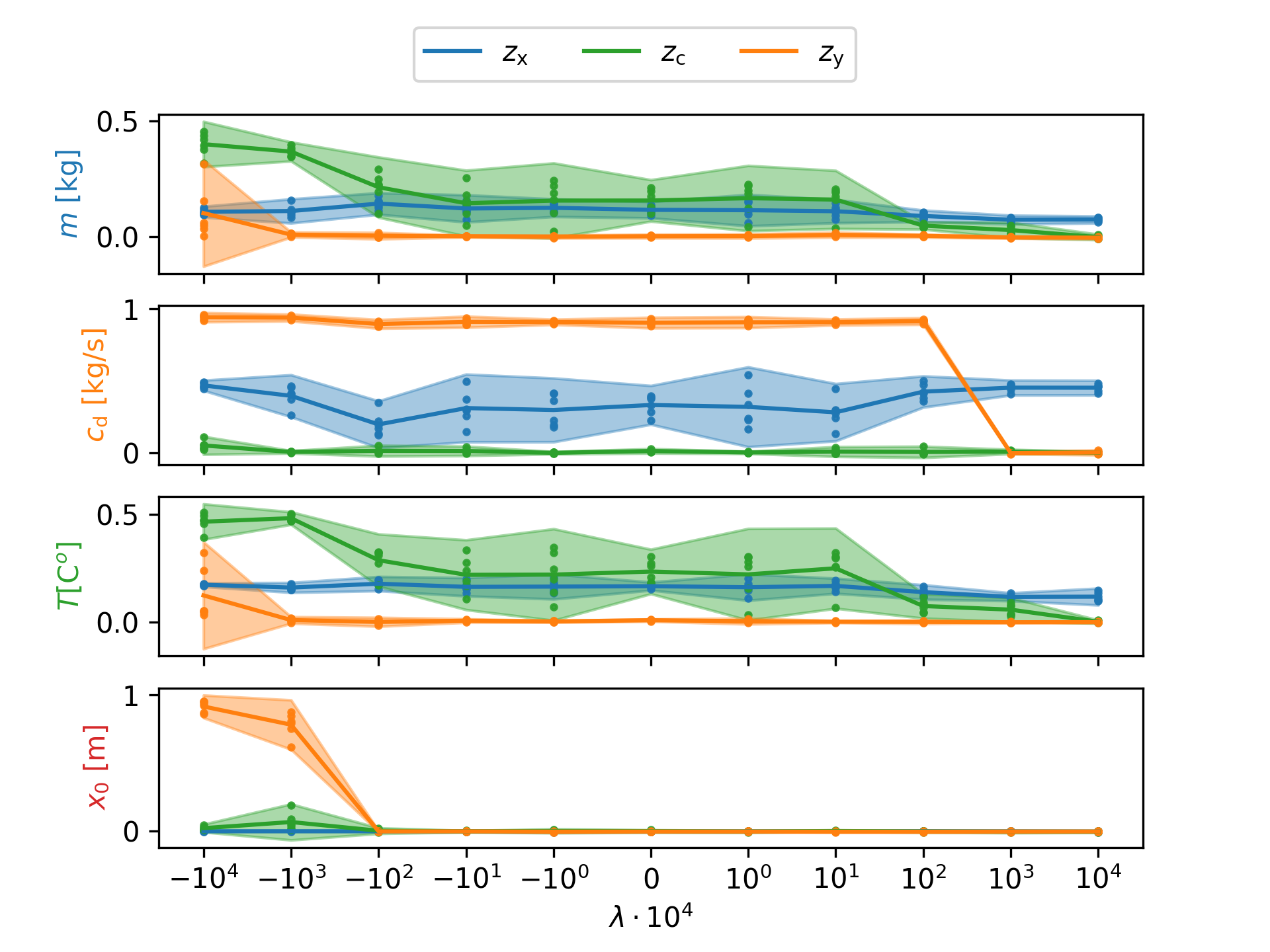}}
	\caption{$R^2$ value per subset of the latent variables and generative factor as a function of $\lambda$, averaged over $6$ runs. The shaded intervals correspond to two standard deviations.}
	\label{fig:oscillator_lambda_study}
\end{figure}

The trade-off between invariance of the domain and class latent variables to non-domain or class influences and prediction accuracy can be adapted by tuning the GRL hyperparameter $\lambda$. A parameter study is performed in order to assess the impact of different choices for $\lambda$ on the learned representation. The model is trained for varying values of $\lambda = \{ -1, -1/10, -1/100, -1/1000, 0, 1/1000, 1/100, 1/10, 1 \}$, and the metric described in Section \ref{section:quantitative_evaluation} is computed for each trained model using linear regression. The training and testing datasets $D$ and $D'$ are composed of $2048$ samples each. To account for the impact of randomness in the synthetic dataset, neural network parameter initialization, and training procedure, the results for each value of $\lambda$ are averaged over multiple runs. The values of the metric for each generative factor and subset of the latent space, as a function of $\lambda$, are shown in \autoref{fig:oscillator_lambda_study}.

It can be seen that the sign of $\lambda$ determines the nature of the training procedure, with positive values resulting in adversarial training. For negative values of $\lambda$ the training becomes collaborative, in the sense that the encoder attempts to find approximate posterior distributions over $\bm{z}_\mathrm{c}$ and $\bm{z}_\mathrm{y}$ that are jointly informative about their respective modality as well as the response measurements $\bm{x}$. The magnitude of $\lambda$ determines the strength of the adversarial or collaborative training. To aid in the interpretation of the results, the behavior of the model is classified into four regimes. When $\lambda$ approaches $-1$ from above, the training is \textit{strongly collaborative}, and the tasks of minimizing the error in the reconstruction of $\bm{x}$ and the prediction of the domain and class variables $\bm{c}$ and $\bm{y}$ are jointly prioritized. This is reflected by the relatively high scores obtained by the subsets $\bm{z}_\mathrm{c}$ and $\bm{z}_\mathrm{y}$ for the generative factors $m$ and $x_\mathrm{0}$. For $\lambda \rightarrow 0^-$, the auxiliary tasks are prioritized over the main task, and the amount of information about $m$ and $x_\mathrm{0}$ that is encoded in $\bm{z}_\mathrm{c}$ and $\bm{z}_\mathrm{y}$ is limited. In this regime the behavior can be characterized as \textit{weakly collaborative}. Conversely, for small positive values of $\lambda$ the training becomes \textit{weakly adversarial}. In this regime the encoder will seek latent codes over $\bm{z}_\mathrm{c}$ and $\bm{z}_\mathrm{y}$ that are uninformative about $\bm{x}$. Further increasing the GRL coefficient such that $\lambda \rightarrow 1$ yields a \textit{strongly adversarial} model, and any information that can be used to reconstruct the domain and class variables is heavily weighted against the potential improvement in the reconstruction of $\bm{x}$. In this case the encoder fails to capture the variation in any of the generative factors.

\subsection{Bridge case study}

\subsubsection{Case study description}
The final synthetic case study utilizes the two-span bridge benchmark presented in \cite{Tatsis2019}, illustrated in \autoref{fig:examples_overview}(c). Members of a homogeneous population \citep{Bull2021} of bridges are subjected to controlled loading tests, where a vehicle with known mass and moving at a constant velocity is used to excite the bridge response. The response is obtained as a strain influence line, expressed in parts per thousand (\textperthousand), measured by a point strain gauge placed at a distance of $5.625$ m from the start of the bridge, and at a height of $0.1$ m from the bottom of the cross section. Each time-series is composed of $64$ measurements, equally spaced in time $t \in [1, 21]$ s, where $t=0$ s is the moment the vehicle enters the bridge. 

The behavior of each bridge is partially determined by the unknown vertical stiffnesses of the supports $k_{\mathrm{v},1}$, $k_{\mathrm{v},2}$ and $k_{\mathrm{v},3}$, which are taken to vary between different bridges due to variability in the design, construction and soil conditions. The boundary conditions are known to be symmetric such that $k_\mathrm{v,1} = k_\mathrm{v,3}$. The base-10 logarithms of the vertical stiffnesses are considered as physics-grounded latent variables. In the horizontal direction, only the left support has a large stiffness, while the rest are unconstrained. It is assumed that the position of the central pier can vary between members of the population by up to $\pm 1.0$ m from $L/2$. Furthermore, fluctuations from the prescribed reference vehicle velocity $v_{\mathrm{ref}} = 1$ m/s were observed during the tests that can not be accounted for in the nominal physics-based model. These fluctuations are modeled as a multiplicative term $\delta_\mathrm{v}$ such that $v = \delta_\mathrm{v} \cdot v_\mathrm{ref}$. Noisy measurements of the vehicle velocity, and the known pier offsets $\delta_\mathrm{s}$ are included as domain variables. It is assumed that the bridge decks are prone to deterioration in a region around the supports. During inspections performed by experts, each bridge is assigned scores $\bm{y} = (y_1, y_2)$, quantifying the deterioration of the structure near the left and middle supports respectively. Each $y_i$ takes values between zero and one, where zero represents pristine condition and unity corresponds to severe damage of the cross-section at that position. These scores are considered as class observables. Finally, a small variability is considered in the vehicle load such that $F = \delta_\mathrm{F} \cdot F_\mathrm{ref}$, where $F_\mathrm{ref} = 100$ kN is the reference load. This variability is caused by deviation in the transverse position of the vehicle, and is considered as an unknown confounding influence. The quantities involved in the case study along with their prior and ground truth distributions are summarized in \autoref{tab:bridge_variables}

\begin{table}[htb!]
\centering
\caption{Summary of physics-based, class and domain variables for the two-span bridge case study.}
\label{tab:bridge_variables}
\begin{tabular}{@{}cccccc@{}}
\toprule
Variable & Unit & Type & Prior distribution & Ground truth & Reference value \\ 
\midrule
$\log_{10} k_{\mathrm{v},1}$ & N/m & Physical & $\mathcal{U}(9.0, 12.0)$ & $\mathcal{U}(9.5, 11.5)$ & 11.5 \\
$\log_{10} k_{\mathrm{v},2}$ & N/m & Physical & $\mathcal{U}(9.0, 12.0)$ & $\mathcal{U}(9.5, 11.5)$ & 11.5 \\
$y_1$ & - & Class & - & $\mathcal{U}(0.0, 1.0)$ & 0.1 \\
$y_2$ & - & Class & - & $\mathcal{U}(0.0, 1.0)$ & 0.1 \\
$\delta_\mathrm{v}$ & - & Domain & - & $\mathcal{U}(0.9, 1.1)$ & 1.0 \\
$\delta_\mathrm{s}$ & m & Domain & - & $\mathcal{U}(-1.0, 1.0)$ & 0.0 \\
$\delta_\mathrm{F}$ & - & Unknown & - & $\mathcal{U}(0.9, 1.1)$ & 1.0 \\
\bottomrule
\end{tabular}
\end{table}

Training data for the full and nominal physical models is generated using the FE model of \cite{Tatsis2019} as a simulator. The FE model is composed of quadrilateral isoparametric plane stress elements with $9$ nodes each, arranged in a $200 \times 6$ grid. The length, width and height are assumed constant and equal to $L = 25.0$ m, $w = 0.1$ m and $h = 0.6$ m for all of the bridges. The material is linear elastic, with Young's modulus $E = 200$ GPa, density $\rho = 7850$ kg/m$^3$ and Poisson's ratio $\nu = 0.3$. All supports are modeled as linear springs in the vertical direction. The equations of motion are integrated from $t_0=0$ seconds to $t_1=25$ seconds with a timestep of $\mathrm{d}t = 0.00025$ seconds, using an implicit Newmark scheme with parameters $\gamma=1/2$ and $\beta=1/6$. The deterioration is modeled as a reduction of the cross section width, ranging from $0\%$ (for $y_i = 0.0$) to $90\%$ (for $y_i = 1.0$) in the region spanning $L/20$ around the corresponding support. It is noted that the deterioration is intentionally exaggerated to a large extent to ensure that it can be observed in the strain influence lines. The damaged regions are illustrated in \autoref{fig:examples_overview}(c). 

The nominal physics-based model is assumed to be a simplistic but representative model for the behavior of the bridges in the population under study in their nominal condition, obtained through the procedure described in Section \ref{section:case_studies}. In addition to the physics-grounded latent variables, the nominal model also includes the parameter $\delta_\mathrm{s}$ as an input, describing the offset of the pier relative to the center of the bridge in the longitudinal direction. Given the vertical stiffness of the abutments and support and the offset of the central pier, the nominal model returns a time-series of strains.

The response, domain and class observables are contaminated with i.i.d. samples of Gaussian white noise with standard deviations $\sigma_\mathrm{x} = \sigma_\mathrm{c} = \sigma_\mathrm{y} = 10^{-4}$. It is important to note that this case study is not intended to be a realistic representation of system identification and SHM for bridges, since it circumvents several important practical difficulties such as ensuring the consistency and alignment of data collected over long time-scales from a large number of structures. Furthermore, it is assumed for simplicity that the degradation condition of the bridges does not change significantly within the amount of time required to obtain the dataset.

\subsubsection{Qualitative assessment of disentanglement}
\begin{figure}[htb!]
    \centering
    \FIG{\includegraphics[width=1.0\textwidth]{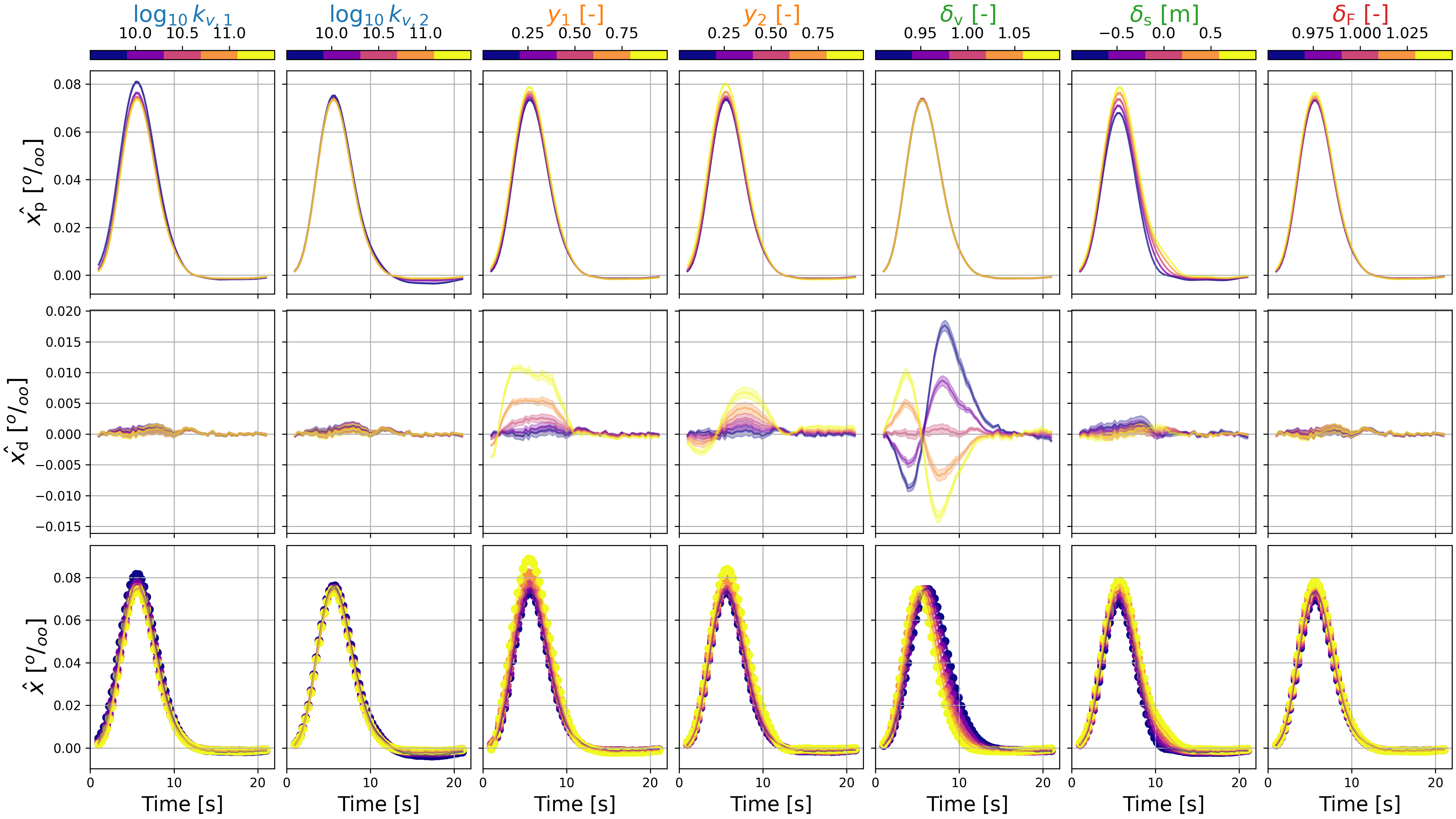}}
    \caption{Mean prediction and $\pm 2\sigma$ uncertainty bounds for the physics-based $\hat{\bm{x}}_\mathrm{p}$ and data-driven $\hat{\bm{x}}_\mathrm{d}$ components, and combined prediction $\hat{\bm{x}}$ while traversing the generative factors. The input response measurements are denoted as dots in the bottom row.}
    \label{fig:bridge_pred_1}
\end{figure}

The model is trained with $\lambda = 1/1024$ and $d_{z_\mathrm{c}} = d_{z_\mathrm{y}} = 4$. The predictions generated by the model while traversing each of the generative factors are shown in \autoref{fig:bridge_pred_1}. It can be seen that the data-driven component of the decoder is prevented from capturing variability in the reconstructed response when varying $\log_{10} k_{\mathrm{v},1}$, $\log_{10} k_{\mathrm{v},2}$ and $\delta_\mathrm{F}$, but is able to contribute to the components caused by the variation of the domain and class generative factors. Furthermore, the figure illustrates that the unknown confounder $\delta_\mathrm{F}$ can be partially accounted for by the physics-based model. This is in contrast to the oscillator example (Section \ref{section:oscillator_example}) where the influence of the unknown confounder could not be accounted for by the known physics. 

\begin{figure}[htb!]
    \centering
    \FIG{\includegraphics[width=1.0\textwidth]{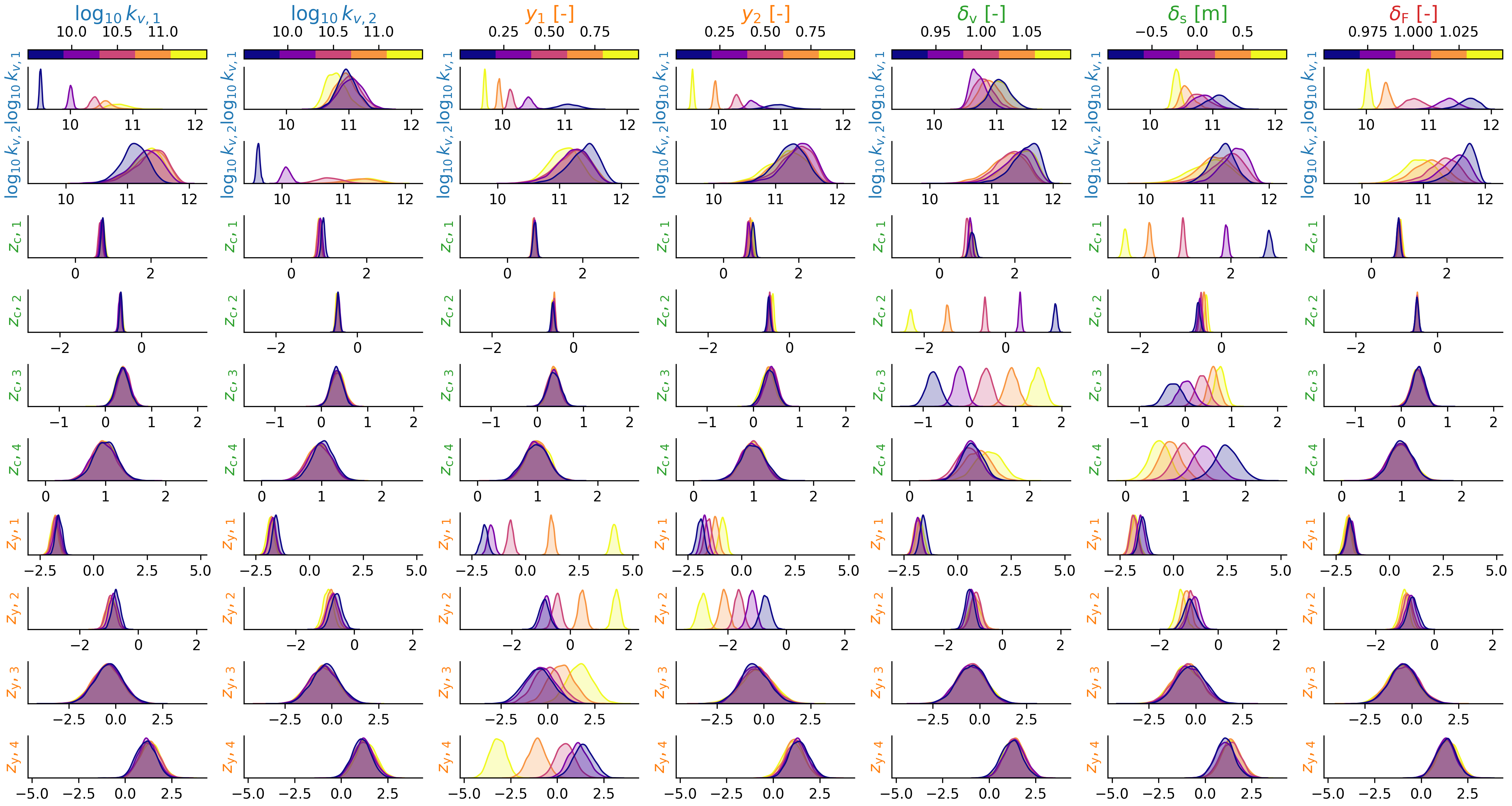}}
    \caption{Visualization of the VAE latent space during traversal of the generative factors. Each column corresponds to variation of a single generative factor, and each row shows the marginal approximate posterior distribution of a single latent variable.}
    \label{fig:bridge_latent_1}
\end{figure}

The previous conclusions are further supported by the traversal of the latent space, shown in \autoref{fig:bridge_latent_1}, which indicates that the domain and class subsets of the latent variables encode information that enables the auxiliary decoders to predict the domain and class labels, and the response decoder to correct the physics-based model prediction. It can be seen that the influence of the unknown confounder $\delta_{\mathrm{F}}$ is partly captured as variability in the physics-based subset $\bm{z}_{\mathrm{x}}$, indicating that model form uncertainty is compensated by inferring an ``effective'' value of the physics-grounded latent variables. \autoref{fig:bridge_latent_1} also suggests that the domain and class latent variables only capture variability in the corresponding generative factors, whereas the physics-grounded latent variables are always active, providing additional evidence for the claim that the adversarial objective induces a disentangled representation while prioritizing the use of the known physics. Importantly, \autoref{fig:bridge_pred_1} and \autoref{fig:bridge_latent_1} illustrate that the adversarial training can feasibly constrain $g_{\bm{\theta}}(\bm{z}_{\mathrm{c}}, \bm{z}_{\mathrm{y}})$, such that it only contributes to the prediction when justified by additional domain and class observables.

\subsubsection{Application to damage identification}
\begin{figure}[htb!]
	\centering
    \begin{subfigure}{1.0\textwidth}
    	\includegraphics[width=1.0\textwidth]{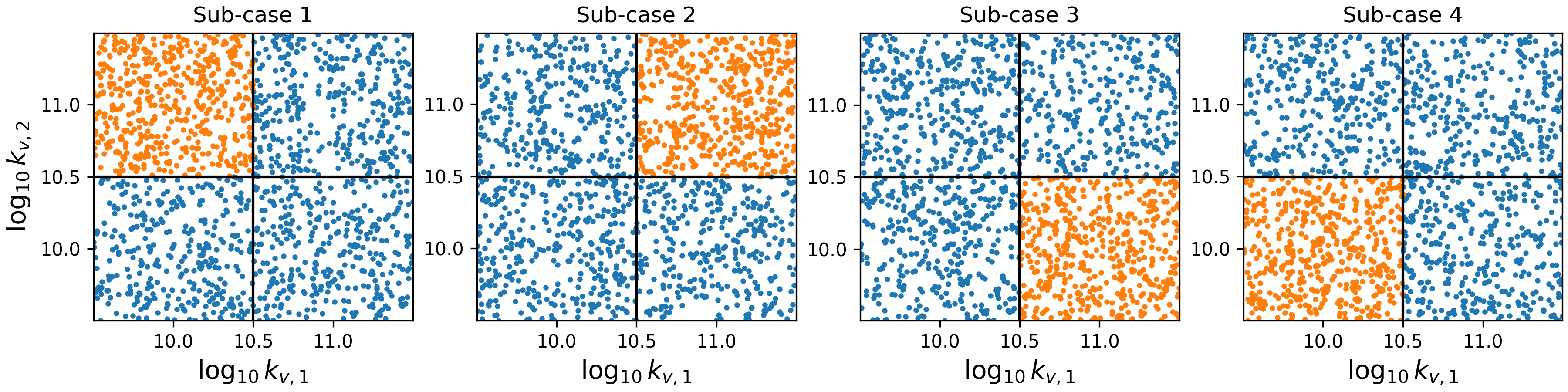}
    	\label{fig:bridge_domain_interpolation}
    \end{subfigure} \\
    \begin{subfigure}{1.0\textwidth}
    	\includegraphics[width=1.0\textwidth]{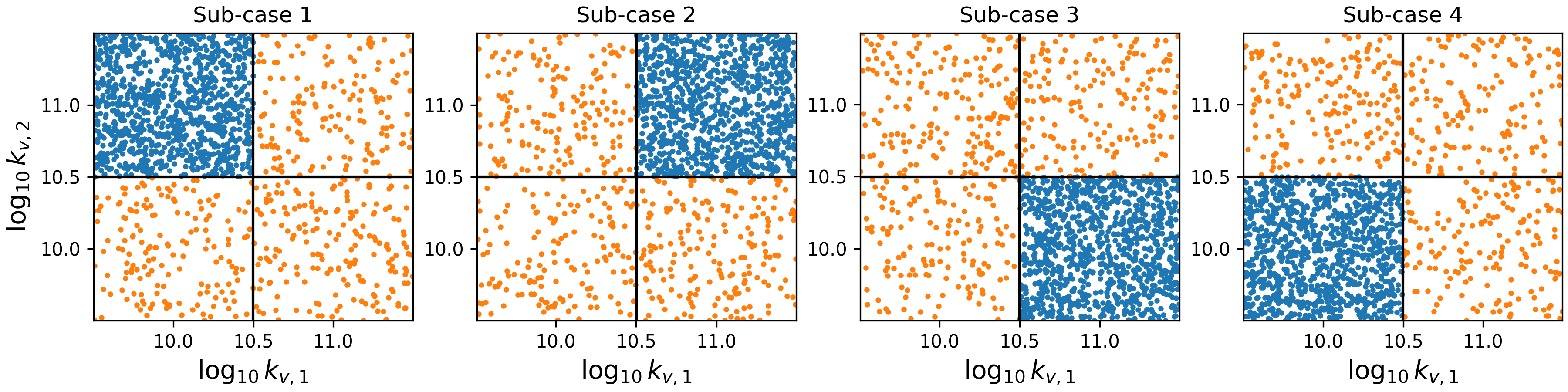}
    	\label{fig:bridge_domain_extrapolation}
    \end{subfigure}
    \caption{Samples of physics-grounded generative factors used for creating the synthetic training set (blue) and test set (orange). Two cases are constructed in order to evaluate performance in interpolation (top) and extrapolation (bottom).}
    \label{fig:example_2_domain}
\end{figure}

As discussed in Section \ref{section:approach}, the model is trained in a fully supervised manner to simultaneously reconstruct the domain and class variables from the input measurements, making it possible to handle tasks such as damage detection, where predicting the class labels $\bm{y}$ from input measurements $\bm{x}$ is of interest. Given a trained model, the condition labels of a similar bridge can be predicted from response measurements. The performance is evaluated in two different cases, illustrated in \autoref{fig:example_2_domain}, referred to as ``interpolation'' and ``extrapolation'' respectively. For each case, the space of physics-grounded generative factors is subdivided into four quarters. In the interpolation case, the model is trained on $N_\mathrm{train} = 1024$ samples from three quarters and evaluated on $N_\mathrm{test} = 512$ samples from the fourth. In the extrapolation case, the model is trained on data from a single quarter and evaluated on the remaining three, using the same train and test set sizes. All other generative factors are sampled from the ground truth distributions presented in \autoref{tab:bridge_variables}. To obtain a more comprehensive evaluation, each of the two cases is divided into four sub-cases, over which the results are averaged. 

The proposed Disentangled Physics-Informed Variational Autoencoder (DPIVAE), using two different hyperparameter settings denoted as DPIVAE-A and DPIVAE-B, is compared with linear regression (LIN), Gaussian Process Regression (GPR) and a Multi-Layer Perceptron (MLP). For DPIVAE-A the GRL is not utilized, i.e. $\lambda = -1$, and separate encoders are used for each subset of the latent variables. For DPIVAE-B the GRL hyperparameter is taken as $\lambda = 1/1024$. The GPR is implemented with a Radial Basis Function kernel and additive Gaussian white noise. The MLP is formulated with two hidden layers, each with a width of $64$ units and a Rectified Linear Unit (ReLU) activation function. Results in terms of interpolation and extrapolation performance of each model is quantified in terms of the R$^2$ and Mean Squared Error (MSE), shown in \autoref{tab:example_2_metrics}.  These results are intended to highlight that the performance of the different models is comparable, and that the proposed model can feasibly be used to predict the class variables in a complex high-dimensional case study. Manual hyperparameter tuning is performed for the models involved in the comparison.

\begin{table}[htb!]
\centering
\caption{Mean and standard deviation of $R^2$ and $\mathrm{MSE}$ for the task of predicting $\bm{y}$, averaged over $6$ runs.}
\label{tab:example_2_metrics}
    \begin{tabular}{lcccc}
    \toprule
      & \multicolumn{2}{c}{Interpolation} & \multicolumn{2}{c}{Extrapolation} \\
    \cline{1-5}
     Model & R$^2$$(\uparrow)$ & MSE$(\downarrow)$ & R$^2$$(\uparrow)$ & MSE$(\downarrow)$ \\
    \midrule
     DPIVAE-A & 0.905 $\pm$ 0.023 & 0.008 $\pm$ 0.002 & 0.676 $\pm$ 0.153 & 0.027 $\pm$ 0.013 \\
     DPIVAE-B & 0.943 $\pm$ 0.012 & 0.005 $\pm$ 0.001 & 0.809 $\pm$ 0.090 & 0.016 $\pm$ 0.008 \\
     GPR      & 0.957 $\pm$ 0.022 & 0.004 $\pm$ 0.002 & 0.820 $\pm$ 0.112 & 0.015 $\pm$ 0.009 \\
     LIN      & 0.863 $\pm$ 0.005 & 0.011 $\pm$ 0.000 & 0.617 $\pm$ 0.285 & 0.032 $\pm$ 0.024 \\
     MLP      & 0.839 $\pm$ 0.035 & 0.013 $\pm$ 0.003 & 0.365 $\pm$ 0.397 & 0.053 $\pm$ 0.033 \\
    \bottomrule
    \end{tabular}
\end{table}

The proposed approach using adversarial training doesn't result in any improvement over existing approaches for the specific interpolation and extrapolation tasks in the present case study. This can be attributed to the adversarial training, which forces the encoder to weigh any information that is relevant to the prediction of $\bm{y}$ against the potential improvement it provides towards the reconstruction of $\bm{x}$. It can be seen that the GPR outperforms all other models while only using a fraction of the parameters, possibly due to the smoothness of the input influence line measurements. The results also indicate that the model performs better when using a single encoder combined with adversarial training in both interpolation and extrapolation. Despite this negative result, we speculate that the disentangled representation induced by the architecture, the invariance of the class latent variables to unknown confounding influences, and the incorporation of the known physics, might be beneficial in certain tasks. Future work will aim to investigate the factors that affect the performance of the proposed approach, and the conditions under which it can provide a benefit in class prediction tasks. This analysis indicates that the proposed model performs on par with other commonly used data-driven models, but with the added benefit of ensuring the proper use of the known physics and the additional interpretability of the physics-grounded latent space.

\section{Discussion}
\label{section:discussion}

\subsection{Contributions and strengths}
The results presented in Section \ref{section:case_studies} indicate that the proposed architecture and adversarial objective effectively constrain the posterior distribution over domain and class latent variables, and by extension, the flexibility of the data-driven decoder components. The constraint is controlled by an interpretable hyperparameter that determines the strength of the gradient reversal. This allows for the main and auxiliary decoders to be trained in a collaborative or adversarial manner. This hyperparameter effectively controls the relative importance of the physics-based and data-driven components, and can be used to encourage the model to preferentially utilize the known physics. When the training is adversarial, the domain and class latent spaces encode features of the response measurements that can be related to the observed domain or and class variables, and that can not be accounted for by the known physics. Simultaneously, the data-driven components of the model are constrained to avoid overriding the physics-based model predictions. Because neither the domain or class observables are necessary during model evaluation, the proposed approach has the potential to reduce the need for cumbersome and expensive data collection methods, such as those involving elaborate experimental procedures or expert assessments. 

\subsection{Assumptions and limitations}
\label{section:assumptions_and_limitations}
It is important to consider the assumptions and limitations of the proposed approach. One of the main drawbacks of the model is the additivity assumption imposed on the physical, domain and class components of the response. It is expected that the model will perform sub-optimally when this assumption is violated. Furthermore, the accuracy of the inferred physics-grounded latent variables will depend on the relative contribution of the physics, domain and class influences to the measured response. Significant domain and class contributions to the response, or violating the additivity assumption of \autoref{eq:hybrid_decoder}, can lead to inaccurate inference of physics-grounded latent variables and large uncertainty in the predictions. Additionally, the model requires that multiple types of data are available, namely measurements of the structural response and information on domain and class. In SHM applications, this might necessitate data alignment procedures of response measurements, environmental conditions and damage level descriptions, and could potentially limit the immediate applicability of the proposed architecture. It is worth mentioning that the interaction between the encoder, decoders, the GRL, and the known physics can be unintuitive in some applications, limiting the applicability of the approach and potentially necessitating implicit supervision by a human expert. 

\subsection{Practical considerations}
Specifying an appropriate value of $\lambda$ for a given learning problem is not straightforward. Schemes for scheduling or adaptively tuning the strength of the GRL during training have been proposed \citep{Qu2025, Li2023, Ganin2015}, but have not been considered in this work. Instead, we focus on providing intuition and clarity regarding the influence of $\lambda$ through the qualitative and quantitative results presented in Section \ref{section:case_studies}. Furthermore, it is known that adversarial training can be unstable \citep{Wiatrak2020}. Throughout this work, occasional instability and overfitting were observed when using small datasets and large batch sizes. Depending on the case study and the value of the $\lambda$ hyperparameter, oscillatory behavior may also occur. We found that these issues could be addressed by adjusting the $\lambda$ hyperparameter, implementing early stopping based on the value of the ELBO on a held-out validation set, and reducing the batch size.

The dimensionality of the latent space is an important design parameter in VAE, and excessively small or large dimensionality can result in poor reconstruction quality \citep{Doersch2021}. Depending on the available computational budget and problem complexity, approaches for determining an appropriate dimensionality are often based on manual trial and error or grid search \citep{Sejnova2024}. More sophisticated approaches include dynamically adjusting the number of latent variables during optimization \citep{DeBoom2021, Sejnova2024}, automatic relevance determination \citep{Saha2025} and multi-stage models \citep{Dai2019}. A key advantage of VAE in engineering, physical, and scientific applications, is that domain knowledge can guide reasoning about the type and number of the dominant generative factors in the data, informing the design of the latent space. VAE are generally insensitive to over-specification of the latent space dimensionality, with superfluous dimensions becoming inactive and ignored by the decoder \citep{Yeung2017, Asperti2019}. Choosing the dimensionality of the domain and class latent space to be a multiple of the expected number of generative factors, based on domain knowledge, and subsequently refining this choice by monitoring the number of inactive dimensions after training can therefore be a viable approach.

It is not possible to provide a rule-of-thumb about the amount of data required for effective training. This would be  dependent on the specific problem, noise levels, physics-based model, and accuracy of the domain and class information, and also on the particular architecture choices (e.g. the number and depth of layers used in the feed-forward NNs in the encoder and decoder). In some applications, the  incorporation of the known physics might lead to a reduction in the data requirements. This has not been investigated in the current paper since it is believed that it would be strongly dependent on the case study chosen, rather than offering any general insight.

\section{Conclusions}
\label{section:conclusions}
The present work contributes to the emerging applications of probabilistic generative models in engineering, by investigating disentangled and invariant representation learning as a tool for grounding VAE to the known physics. Specifically, a Physics-Enhanced Machine Learning strategy utilizing a VAE architecture is proposed, with the aim of learning a disentangled representation of physical, domain and class confounding influences that are present in the response measurements of physical systems. This is achieved by having the decoder and latent space of the VAE be semantically and functionally separated into data-driven and physics-grounded branches. An easy to implement regularization method based on the GRL is used to constrain the data-driven components, resulting in a model that preferentially utilizes the known physics. An interpretable and intuitive hyperparameter is used to specify the strength of GRL, and wether the model is trained in a collaborative or adversarial manner. Moreover, a strategy for quantifying the type and relative amount of information encoded in different sets of latent variables is proposed, yielding insights on the degree of disentanglement achieved by the model.

Three synthetic case studies involving a beam, an oscillator, and a population of bridges were investigated. In these cases, a nominal model representing the partially known physics was available or built from a simulator. For each case, noisy observations of the structural response and information on domain (the environmental and operational conditions that a system is exposed to) and class (the characteristics of a structure related to the existence and extent of damage and degradation) are assumed available. It was shown that the proposed architecture promotes the learning of disentangled representations, and mitigates the issues that occur when including physics-based components in standard VAE. Furthermore, it was shown that the proposed approach is able to: (i) Preferentially utilize the known physics, resulting in an interpretable and physically meaningful posterior distribution over physics-grounded latent variables, (ii) Accurately reconstruct the structural response in the presence of domain and class influences that are not described by the known physics, and (iii) Predict the class variables associated with a structure under previously unseen conditions using noisy measurements of the structural response. 

Although the results of the case studies do not indicate improvement in the prediction of class variables, compared to commonly used data-driven approaches, it is likely that the invariance of the learned domain and class representations with respect to unknown confounding influences can be advantageous for certain problems. Future work will aim to investigate this, as well as the performance of the approach in more complex tasks and in real-world problems. Other possible avenues for future work include the extension of the approach to the semi-supervised setting, the application to dynamical systems described by ordinary differential equations, and automating the tuning of the GRL hyperparameter. 

\begin{Backmatter}

\paragraph{Funding Statement}
This publication is part of the project LiveQuay: Live Insights for Bridges and Quay walls (project number NWA.1431.20.002) of the research programme NWA UrbiQuay which is (partly) funded by the Dutch Research Council (NWO).

\paragraph{Competing Interests}
None.

\paragraph{Data Availability Statement}
The code and data (generated via the synthetic use cases) needed to replicate the results shown in this paper can be accessed through the link: \url{https://doi.org/10.5281/zenodo.15813028} \citep{Koune2025}.

\paragraph{Ethical Standards}
The research meets all ethical guidelines, including adherence to the legal requirements of the study country.

\paragraph{Author Contributions}

Conceptualization: I.K.; A.C. Methodology: I.K; A.C. Data curation: I.K.; A.C. Data visualisation: I.K.; A.C. Writing original draft: I.K; A.C. All authors approved the final submitted draft.

\begin{appendix}
\section{Implementation details}
\label{appendix:appendix_A}

\paragraph{Encoder formulation}
The encoder reduces the dimensionality of the input measurements and maps them to vectors of mean values $\bm{\mu}_\mathrm{\phi}(\bm{x})$, standard deviations $\bm{\sigma}_\mathrm{\phi}(\bm{x})$ and a lower triangular matrix $\bm{L}'_\mathrm{\phi}(\bm{x})$. The lower triangular factor $\bm{L}_\mathrm{\phi}(\bm{x}) = \bm{L}'_\mathrm{\phi}(x) + \bm{\sigma}_\mathrm{\phi}(\bm{x}) \bm{I}$ is the Cholesky decomposition factor of the covariance matrix $\bm{\Sigma}_\mathrm{\phi}(\bm{x})$, i.e. $\bm{\Sigma}_\mathrm{\phi}(\bm{x}) = \bm{L_\mathrm{\phi}(\bm{x}) L_\mathrm{\phi}(\bm{x})}^T$, such that the posterior distribution corresponding to each input is a multivariate Normal distribution. The encoder outputs the log of the standard deviations, which are then exponentiated to avoid negative values. The reparametrization trick \citep{Kingma2022} is exploited to define the latent variables $\bm{z}$ as a deterministic transformation of a noise variable $\bm{\epsilon} \sim p(\bm{\epsilon})$. This facilitates the computation of unbiased Monte Carlo gradient estimates of the objective with respect to the variational parameters, using automatic differentiation. With the exception of the introductory examples presented in Section \ref{section:challenges}, the encoder is everywhere formulated as a single feed-forward NN using a shallow architecture with a single hidden layer. The input and hidden layer widths are $[d_\mathrm{x}, 128]$, where $d_{\mathrm{x}}$ is the dimensionality of the input. The output layer is composed of three heads, corresponding to the mean, standard deviation and covariance outputs. The mean and standard deviation heads have output sizes of $d_\mathrm{z}$, while the covariance head has an output size of $d_{\mathrm{z}}^2$, where $d_\mathrm{z} = d_\mathrm{z_\mathrm{x}} + d_\mathrm{z_\mathrm{c}} + d_\mathrm{z_\mathrm{y}}$, with $d_\mathrm{z_\mathrm{i}}$ denoting the size of the $i$'th subset of the latent space. A ReLU activation function is applied on all layers except the final output layer. For the introductory examples, an independent encoder network is used for each subset of the latent variables. The hidden layer widths of each independent network are set to 64 units, and the output shapes are adjusted according to the dimensionality of the corresponding subset of the latent variables. To further ensure numerical stability, the outputs of all encoder NNs are clamped within ranges of values that are expected to be encountered for the case studies investigated in this work. 

\paragraph{Decoders}
The decoder of the response is formulated as a feed-forward NN with a single, 128-unit-wide hidden layer and a ReLU nonlinearity at the output of the hidden layer. The size of the input is $d_{\mathrm{z}_{\mathrm{c}}} + d_{\mathrm{z}_{\mathrm{y}}}$ and the output size is $d_{\mathrm{x}}$. A gradient reversal layer is placed at the input of this network. For the structural response prediction, the standard deviation $\sigma_\mathrm{x}$ is included in the vector $\boldsymbol{\theta}$ and jointly optimized with the NN hyperparameters. The auxiliary networks are formulated with a single hidden layer with a width of $64$ units and a ReLU nonlinearity between the input and the hidden layer. The auxiliary decoders are composed of two prediction heads, responsible for the mean prediction and standard deviation respectively. The input and output shapes are $d_{\mathrm{z}_{i}}$ and $2 \cdot d_i$, where $i$ denotes the corresponding domain or class modality.

\paragraph{Conditional prior networks}
The conditional prior distributions are formulated as factorized Gaussian distributions. The corresponding neural networks use a single hidden layer with a width of $64$ units. The input and output shapes are also adjusted to $d_\mathrm{i}$ and $d_{\mathrm{z}_i}$ respectively, where the subscript $i \in \{x, c, y \}$ denotes the corresponding modality and subset of the latent space. 

\paragraph{Latent variable transformation}
To facilitate the application of the model to cases involving physics-grounded latent variables with bounded support, and to improve numerical stability, all parameters are transformed from an unbounded and normalized base latent space to the target latent space in which they are defined. This is achieved by applying a sequence of deterministic transformations to the samples and corresponding scaling of the log-densities. In the following, variables in the base space are denoted as $u$. The samples at the output of the encoder are first bounded by applying the logistic transform $u' = \frac{1}{1+e^{-u}}$, and subsequently scaled and shifted using an affine transform $z = u' \cdot (\mathrm{UB} - \mathrm{LB}) + \mathrm{LB}$ to bound the variables to their specified supports defined by the lower and upper bound $\mathrm{LB}$ and $\mathrm{UB}$. The samples and densities can also be mapped from the target latent space to the base latent space by applying the corresponding inverse transforms in reverse order.

\paragraph{Optimization}
Optimization is carried out using the Adam algorithm \citep{Kingma2017} with minibatch gradient estimation \citep{Kingma2022}. The model is trained for up to $20,000$ iterations with a batch size of $64$. Early stopping is implemented by monitoring the value of the ELBO, evaluated on a held-out validation set with size $N_\mathrm{val} = 512$. The training is terminated if no improvement of the ELBO is observed over $2,000$ iterations. Gradient and objective estimates are obtained using $16$ Monte Carlo samples during training, $64$ during validation, and $512$ during evaluation, although in practice the training was found to be insensitive to the number of samples. All learning rates are set to $0.001$, except for the learning rate of the standard deviation parameter for the response $\sigma_\mathrm{x}$, which is set to $0.005$. The $\alpha$ and $\beta$ hyperparameters of the optimization objective are taken as $\beta = \alpha_\mathrm{x} = \alpha_\mathrm{c} = \alpha_\mathrm{y} = 1.0$ for all the experiments presented in this work.

\paragraph{Visualization}
Figures illustrating the traversal of the latent space and the space of reconstructions are provided for each case study. Samples from the latent space and reconstructions are obtained as follows: Five linearly spaced values between the $1^\mathrm{st}$ and $99^\mathrm{th}$ percentile of the ground truth distribution are computed for each generative factor in turn, while the remaining generative factors are fixed to a constant value. For each combination of generative factors, $1000$ realizations of response measurements are generated using the procedure described in Section \ref{section:case_studies}. The model is evaluated on the response measurements, and a single sample is drawn from the approximate posterior distribution for each response measurement. The decoder is then evaluated on each sample from the posterior, yielding deterministic predictions $\hat{\bm{x}}_\mathrm{p}$ and $\hat{\bm{x}}_\mathrm{d}$ from the physics-based and data-driven components respectively. The combined prediction is sampled from $\mathcal{N}(\hat{\bm{x}}_\mathrm{p} + \hat{\bm{x}}_\mathrm{d}, \sigma^2_\mathrm{x} \bm{I})$. The visualizations of the latent space and reconstructions therefore also include the randomness in the data generating process, in addition to the randomness in the approximate posterior distribution and decoder.




\end{appendix}

\printbibliography

\end{Backmatter}

\end{document}